\documentclass[11pt]{article}
 \usepackage{amsmath}
\usepackage{amssymb}
\usepackage{amsthm}

\usepackage{graphicx}
\usepackage{color}
\usepackage{multicol}
\usepackage{mathrsfs}
\usepackage{float}
\usepackage[caption = false]{subfig} 
\usepackage{dsfont}
\usepackage{times}
\usepackage{commath} 
\usepackage{natbib}

\usepackage[linesnumbered,ruled]{algorithm2e}

\newcommand{\be}{\begin{equation}}
\newcommand{\ee}{\end{equation}}
\newcommand{\bes}{\begin{equation*}}
\newcommand{\ees}{\end{equation*}}
\newcommand{\beqn}{\begin{eqnarray}}
\newcommand{\eeqn}{\end{eqnarray}}
\newcommand{\beqns}{\begin{eqnarray*}}
\newcommand{\eeqns}{\end{eqnarray*}}

\newcommand{\lkr}{\left(}
\newcommand{\lkv}{\left[}
\newcommand{\rkv}{\right]}
\newcommand{\rkr}{\right)}
\newcommand{\lfi}{\left\{}
\newcommand{\rfi}{\right\}}
\newcommand{\di}{\displaystyle}

\long\def\ignore#1{}

\newcommand{\bm}[1]{{\mbox{\mathversion{bold}$#1$}}}

\newcommand{\R}{\mathbb{R}}

\newcommand{\om}{\omega}
\newcommand{\lam}{\lambda}

\newcommand{\sig}{\sigma}

\newcommand{\Om}{\Omega}

\newcommand{\EE}{\ensuremath{{\mathbb E}}}

\newcommand{\PP}{\ensuremath{{\mathbb P}}}

\newcommand{\RR}{\ensuremath{{\mathbb R}}}

\newcommand{\card}{\mbox{card}}

\newcommand{\diag}{\mbox{diag}}

\newcommand{\Tr}{\mbox{Tr}}

\newcommand{\vect}{\mbox{vec}}
\newcommand{\Pen}{\mbox{Pen}}
\newcommand{\Err}{\mbox{Err}}

\newtheorem{theorem}{Theorem}
\newtheorem{lemma}{Lemma}

\newtheorem{remark}{Remark}

\newcommand{\calD}{{\mathcal{D}}}
\newcommand{\calG}{{\mathcal G}}
\newcommand{\calJ}{{\mathcal{J}}}

\newcommand{\calM}{{\mathcal{M}}}

\newcommand{\scrP}{\mathscr{P}}

\newcommand{\tk}{\tilde{k}}
\newcommand{\tj}{\tilde{j}}
\newcommand{\tl}{\tilde{l}}
\newcommand{\tp}{\tilde{p}}

 \newcommand{\oPen}{\overline{\Pen}}

\newcommand{\hC}{\widehat{C}}
\newcommand{\hZ}{\widehat{Z}}
\newcommand{\hThe}{\widehat{\Theta}}
\newcommand{\hK}{\hat{K}}
\newcommand{\hL}{\hat{L}}

\newcommand{\tH}{\tilde{H}}

\newcommand{\scrPZC}{\mathscr{P}_{Z,C}}

\long\def\ignore#1{}

\begin{document}

 \title{{The Hierarchy of Block Models }}

\author{{\bf Majid Noroozi}\\
Department of Mathematics and Statistics\\ 
Washington University in St. Louis\\
St. Louis, MO 63130\\ USA \\
{\it majid.noroozi@wustl.edu}
\and
{\bf Marianna Pensky} \\
Department of Mathematics\\
University of Central Florida \\
Orlando, FL 32816 \\
USA \\
{\it Marianna.Pensky@ucf.edu}
}

\date{}

\maketitle

\begin{abstract}
There exist various types of network block  models such as  the Stochastic Block Model (SBM), 
the Degree Corrected Block Model (DCBM), and the Popularity Adjusted Block Model (PABM).
While this leads to a variety of choices, the block models do not have a nested structure. 
In addition, there is a substantial jump in the number of parameters from the DCBM  to the PABM.  
The objective of this paper is formulation of a hierarchy of  block model which does not rely on 
arbitrary identifiability conditions. We propose a Nested Block Model (NBM) that treats the SBM, the DCBM and the PABM 
as its particular cases  with specific parameter values,
and, in addition, allows a multitude of versions that 
are more complicated than DCBM but have fewer unknown parameters than the PABM. 
The latter allows one to carry out clustering and estimation  without preliminary testing, to see
which block model is really true.

\vspace{2mm} 

\vspace{2mm}
{\bf AMS (2000) Subject Classification}: {Primary:  62F12, 62H30.   Secondary:  05C80  }
\end{abstract}

\section*{Funding:} Both authors of the paper were  partially supported by National Science Foundation (NSF)  grants DMS-1712977 and DMS-2014928
 


\section{Introduction}
\label{sec:intro}

Consider an undirected network with $n$ nodes, no self-loops and multiple edges. 
Let $A \in \{0,1\}^{n\times n}$ be a symmetric adjacency  matrix of the network with
$A_{i,j}=1$ if there is a connection between nodes $i$ and $j$,  and $A_{i,j}=0$
otherwise. We assume that 
%
 $A_{i,j}\sim \mbox{Bernoulli}(P_{i,j}),$  
 $1 \leq i < j \leq n,$
where  $A_{i,j}$  are conditionally independent 
given $P_{i,j}$, and  $A_{i,j} = A_{j,i}$, $P_{i,j} = P_{j,i}$ for $i>j$.
The probability matrix  $P$ has low complexity and can be described by a variety of models. 
One of the ways to address this phenomenon is to assume that all nodes in the network can be partitioned 
into {\it communities}, which are groups that exhibit somewhat similar behavior.

The classical  
\cite{erdos59a}  random graph model    assumes that the edges in a random graph are
drawn independently with  an equal probability and  does not allow community structure. 
The simplest  random graph model for networks with community 
structure is the Stochastic Block Model (SBM)  studied by, e.g.,    
\cite{doi:10.1080/0022250X.1971.9989788} and  \cite{JMLR:v18:16-480}. 
Under the $K$-block SBM,  all nodes are partitioned into communities $\calG_k$, $k=1, \ldots, K$, 
and the probability of connection between nodes is completely defined by the communities to which they 
belong: $P_{i,j} = B_{z(i),z(j)}$  where $B_{k,l}$ is the probability of connection between communities 
$k$ and $l$, and $z: \{1,...,n\} \to \{1,...,K\}$  is a clustering function such that $z(i)=k$ whenever $i \in \calG_k$.
The Erd\H{o}s-R\'{e}nyi model   can be  viewed as the SBM with only one community $K=1$.

Since the real-life networks usually contain a very small 
number of high-degree nodes 
while the rest of the nodes have very  low degrees, 
the SBM  fails to explain the 
structure of many networks that occur  in practice. The  Degree-Corrected Block Model (DCBM), 
introduced by Karrer and Newman (2011), 
addresses this deficiency by  allowing these probabilities to be 
multiplied by the node-dependent weights. 
Under the DCBM, the elements of matrix $P$ are modeled as 
\be \label{eq:DCBM}
 P_{i,j }= h_i B_{z(i),z(j)} h_j, \quad i,j = 1, \ldots, n, 
\ee
where $h=[h_1, h_2, ..., h_n]$ is a vector of  the degree parameters of the   nodes, and $B$ 
is the $(K \times K)$ matrix of baseline interaction between communities.  Matrix $B$ and vector  
$h$  in \eqref{eq:DCBM} are defined up to a scalar factor, which is usually fixed via the so called 
{\it identifiability} condition,  that can be 
imposed in a variety of ways. For example, 
\cite{Karrer2011StochasticBA} 
enforce  a constraint of the form  
\be \label{eq:DCBM_ident_con}
\sum_{i\in \calG_k} h_i = 1, \quad     k = 1,...,K.
\ee
 
The DCBM implies that the probability of connection of a node is uniformly proportional 
to the degree of this node  across all communities. This assumption, however,  is violated 
in a variety of practical applications. For this reason, 
\cite{RePEc:bla:jorssb:v:80:y:2018:i:2:p:365-386} 
introduced the Popularity Adjusted  Block Model (PABM). The PABM presents the probability 
of a connection between nodes  as a product of popularity parameters, that depend on the 
communities to which the nodes belong as well as on the pair of nodes themselves:
\begin{equation} \label{eq:PABM-model}
P_{i,j}= V_{i,z(j)} V_{j,z(i)}. 
\end{equation}
Although the  popularity parameters in \eqref{eq:PABM-model}  are defined up to  scalar constants and 
require an identifiability condition  for their recovery, clustering of the nodes and fitting the   
matrix of connection probabilities  do not require  any  constraints. According to \cite{2019arXiv190200431N}, 
if one re-arranged the nodes, so that  the nodes in every community are grouped together, then    
matrix $P$ of the connection probabilities would appear as $(K \times K)$ block matrix 
with every block $P^{(k,l)}$, $k,l = 1, \ldots, K,$ being of rank one.

Having   several types of  block models introduces   a variety of choices, but also 
  leads   to some significant drawbacks. 
Specifically, although  the block models can be viewed as 
progressively more elaborate with the 
Erd\H{o}s-R\'{e}nyi   model  being the  simplest and the PABM being the most 
complex, the simpler models cannot be viewed as  particular cases of the more sophisticated ones as one paradigm. 
For this reason, majority of authors carry out estimation and clustering under the assumption that 
the model which they use is  indeed the correct one.
There are only very few papers that study goodness of fit in block 
models setting, and majority of them are concerned with either testing that there 
are no distinct communities, that is   $K=1$ in SBM or DCBM  
(see, e.g., \cite{banerjee2017optimal},  \cite{gao2017testing} and \cite{pmlr-v80-jin18b}),
or testing the exact number of communities $K=K_0$ in the SBM 
(see, e.g., \cite{gangrade2018testing}, 
 \cite{lei2016} and   \cite{mukherjee2017testing}).  
To the best of our knowledge, \cite{mukherjee2017testing} is the only paper where
testing the SBM versus the DCBM is implemented, and    
the testing in their paper  is carried out  under rather restrictive  assumptions.
%
On the other hand, using the most flexible model, the PABM, may not always be the right choice 
since there is a substantial jump in complexity from the DCBM with   $O(n + K^2)$ parameters to the PABM 
with $O(nK)$ parameters.

The objective of the present paper is to provide a unified approach to block models. 
We would like to point out that we are building a {\it  hierarchy   of block models}, and not a   hierarchical stochastic block model.
In our paper, we consider a multitude of block models and provide an enveloping nested model that includes them all as particular cases.
In what follows, we shall deal only with the graphs where each node belongs to one and only one community, 
thus, leaving aside the mixed membership models studied by, e.g.,  \cite{Airoldi:2008:MMS:1390681.1442798} and \cite{jin2017estimating}. 
Specifically, our purpose is formulation of a hierarchy of  block models which does not rely on arbitrary identifiability conditions, 
treats the SBM, the DCBM and the PABM as its particular cases (with specific parameter values)  
and, in addition, allows a multitude of versions that 
are more complicated than DCBM but have fewer unknown parameters than the PABM. 
The aim of this construction is to  treat all block models as a part of one paradigm and, therefore,  
carry out estimation and clustering without preliminary testing to see which block model fits data at hand.



\section{The hierarchy of block models}
\label{sec:hierarchy}

Consider an undirected network with $n$ nodes that are partitioned into $K$ communities $\calG_k$, $k=1, \ldots, K$, 
by a clustering function $z: \{1,\ldots,n\} \to \{1,\ldots,K\}$ with the corresponding clustering matrix $Z$.
Denote by $B$     the matrix of average connection probabilities between communities, so that
for $k,l = 1,2, \cdots , K,$  one has
\be  \label{eq:B_def} 
  B_{k,l}=\frac{1}{n_k \, n_l} \sum_{i,j = 1}^n P_{ij}\, I(z(i)=k)\, I(z(j)=l),
\ee
where  $n_{k}$ is the number of nodes in the community $k$.

In order to better understand the relationships between various block  models, 
consider a rearranged version $P(Z)$ of matrix $P$ where 
its  first $n_{1}$ rows correspond to  nodes from class~1, 
the next $n_{2}$ rows correspond to  nodes from class~2, and the last $n_K$ rows correspond to  nodes from class~$K$. 
Denote the $(k_1,k_2)$-th block of matrix $P(Z)$ by $P^{(k_1,k_2)} (Z)$. 
Then, the block models vary by how dissimilar matrices $P^{(k_1,k_2)}(Z)$ are. 
Indeed, under the SBM
\be  \label{eq:P_blocks_SBM}  
P^{(k_1,k_2)} (Z) = B_{k_1,k_2} 1_{n_{k_1}} 1_{n_{k_2}}^T
\ee
where $1_k$ is the $k$-dimensional column vector with all elements equal to one.
In the DCBM, there exists a vector $h \in \RR_+^n$, with sub-vectors $h^{(k)} \in \RR_+^{n_{k}}$, $k=1, \ldots, K$, 
such that, for $k_1,k_2 = 1,2, \cdots , K,$
\be  \label{eq:P_blocks_DCBM}  
P^{(k_1,k_2)}(Z) = B_{k_1,k_2} h^{(k_1)} (h^{(k_2)})^T.
\ee
In the PABM, instead of one vector $h$, there are $K$ vectors $\Lambda^{(1)}, \cdots ,\Lambda^{(K)}$ with sub-vectors
\be  \label{eq:Lambda_blocks}  
\Lambda^{(k_1,k_2)} \in \RR_+^{n_{k_1}},  \quad k_1,k_2 = 1,2, \cdots , K.
\ee
In this case, vectors $\Lambda^{(k)}$ form the $(n \times K)$ matrix $\Lambda$ with 
columns partitioned into sub-columns $\Lambda^{(k_1,k_2)}$, and 
\be  \label{eq:P_blocks_PABM} 
P^{(k_1,k_2)}(Z) = B_{k_1,k_2} \Lambda^{(k_1,k_2)} (\Lambda^{(k_2,k_1)})^T,  
\ee
for every $k_1, k_2 = 1,2, \cdots , K$.
Hence, \eqref{eq:P_blocks_SBM} and \eqref{eq:P_blocks_DCBM} coincide if $h \equiv 1_n$, and 
\eqref{eq:P_blocks_PABM} reduces to  \eqref{eq:P_blocks_DCBM} if all columns of matrix $\Lambda$ are identical, i.e.
\be  \label{eq:Lambda_equiv_h_blocks}  
\Lambda^{(k_1,k_2)} \equiv h^{(k_1)},  \quad k_1, k_2 = 1,2, \cdots , K.
\ee
Since  in the DCBM there is only one vector $h$ that models heterogeneity in probabilities of connections,
the ratios $P_{i_1,j}/P_{i_2,j}$ of the  probabilities of connections of two nodes, $i_1$ and $i_2$, 
that belong to the same community, are determined entirely by the nodes  $i_1$ and $i_2$ and are independent 
of the community with which those nodes interact.  
On the other hand, for the PABM, each node has a different degree of popularity (interaction level) with 
respect to  every other community, so that   $P_{i_1,j_1}/P_{i_2,j_1} \neq P_{i_1,j_2}/P_{i_2,j_2}$  
if nodes $j_1$ and $j_2$ belong to different communities.
In the PABM,  those variable popularities are described by the matrix  
$\Lambda \in [0,1]^{n \times K}$
which reduces to a single vector $h$ in the case of the DCBM.  
One can easily imagine the situation where nodes do not exhibit different levels 
of activity with respect to every community but rather with respect 
to some groups of communities, 
``{\it  meta-communities}'', so that there are $L$, $1 \le L \le K$,  
different vectors $H^{(l)} \in \RR_+^{n},  \quad l = 1,2, \cdots , L,$
and each of columns $\Lambda_k$, $k = 1,2, \cdots , K$, of matrix $\Lambda$ is equal to one of vectors $H^{(l)}$.
In other words, there exists a clustering  function $c:\{1,...,K\} \to \{1,...,L\}$ 
with the corresponding clustering matrix $C$  such that
\bes  
\Lambda_k = H^{(l)},\ l=c(k), \  l=1, \ldots ,L,\ k = 1, \ldots , K.
\ees
We   name the resulting model the {\it Nested  Block Model } (NBM)  to emphasize that
the model is equipped with the nested structure that allows to obtain a multitude of popular block models as its particular cases.


\section{The Nested Stochastic Block Model (NBM)}
\label{sec:HBM}



The NBM    contains two types of communities, the regular communities that can be 
distinguished by the average probabilities of connections between them (like in the SBM or the DCBM) 
and the meta-communities that are described by the distinct patterns of probabilities of connections  
of individual nodes across the communities.

Observe that both concepts are quite natural. 
Indeed, in  random network models, specifically in assortative network models that are most common, 
communities are usually loosely defined as groups of nodes that have higher probability of connection than the rest.
In our case, we retain the notion and define communities as groups of nodes with a specific (average) probability 
of connection between them. The meta-communities  refer to the node-to-community
specific connection weights. In DCBM, each node has only one specific weight to account for the difference in 
connection probabilities with the rest; in PABM, the weights can be different for any node and community pair. In our nested model,
we allow some of the node-to-community interactions have the same patterns for a group of communities, the meta-communities.

For instance, consider an example of College of Sciences of a university that includes Departments of 
Biology, Chemistry, Mathematics, Physics, Psychology, Sociology, and Statistics.
Each of the departments forms a natural community, and the average density of connections is much higher within 
the communities than between them. However, one can naturally partition College of Sciences into meta-communities of 
natural sciences (Biology, Chemistry, and Physics), social sciences  (Psychology and Sociology)  and 
mathematics and data sciences (Mathematics and  Statistics). For example, faculty in mathematics and data sciences 
who are working on  various problems in astronomy, genetics, dynamical systems, or theory of chemical reactions 
will have high probability of connections with the natural sciences meta-community. On the other hand, those who
are involved  in, for example, studying  social interactions, 
monitoring cyber and homeland security, or  relationships   between   countries,
will have more dense interactions with the social sciences meta-community. While one can use the PABM and model
interactions between each pair of the departments separately, the patterns within meta-communities may be similar enough 
that  using the  most complex PABM  with $O(nK)$ parameters may not be justified.

Note that  the meta-communities introduced in this paper should not be mixed with the mega-communities 
considered  in \cite{Wakita:2007:FCS:1242572.1242805} and    \cite{li2018hierarchical}. 
The difference between the present paper and the above cited publications is that in 
\cite{Wakita:2007:FCS:1242572.1242805} and  \cite{li2018hierarchical} the  mega-communities are determined by 
intermediate results of the clustering algorithms while we define the meta-communities on the basis of  the distinct 
patterns of the  connection probabilities of nodes with respect to different communities. 
Our approach is also very different from the hierarchical stochastic block model  studied in, 
e.g.,  \cite{li2018hierarchical} or \cite{7769223}. In those papers, the authors 
examine SBMs with a large number of communities, that can be partitioned into groups based on some similarities in the matrix 
of block probabilities. We, on the other hand, deal with  more diverse block models, for which the SBM is the simplest one. 
In addition, the authors in \cite{7769223} impose  assumptions that require the SBM to be very strongly assortative. 
Hence, the only common feature between our paper   and the above mentioned ones is that there exist groups of communities; 
everything else is completely different.

For any $M$ and $K \leq M$, denote by $\calM_{M,K}$ the collection of  all clustering  matrices 
$Z \in \{0,1\}^{M\times K}$  with the corresponding clustering 
function $z: \{1,\ldots,M\} \to \{1,\ldots,K\}$
 such that $Z_{i,k}=1$ iff $z(i) = k$, $i=1, \ldots, M$. Then,  
$Z^T Z = \diag (n_1, \ldots, n_K)$ where  $n_k$ is the size of community $k$, $k=1, \ldots, K$.  
The NBM, with $K$ communities and $L \leq K$ meta-communities, is defined by 
two clustering matrices $Z \in \calM_{n,K}$ and $C \in \calM_{K,L}$  with corresponding clustering 
functions  $z$ and $c$  that, respectively,  partition the $n$ nodes into   $K$ communities, and   
$K$ communities into $L$ meta-communities.
If the $l$-th meta-community consists of $K_l$ communities  and the community sizes
are $n_k$, then the total number of nodes in  meta-community $l$  is $N_l$, where
\be    \label{eq:num_nodes_in_mega} 
 \begin{aligned}
 N_{l}   =    \sum_{k = 1}^K n_k\,  I(c(k)=l), \quad  \sum_{l = 1}^L K_l =K, \quad \sum_{l = 1}^L N_l =n, \quad   l= 1, \cdots , L.  
\end{aligned}
\ee 
The communities are characterized by their average connection probability matrix, with elements $B_{k_1,k_2}$,  
$k_1,k_2 = 1,2, \ldots , K$, defined in \eqref{eq:B_def}.
In order to better understand the meta-communities, consider a   permutation matrix
$\scrPZC$ that arranges nodes into communities consecutively, and orders communities
so  that  the $K_l$  blocks within the $l$-th meta-community are consecutive, $l = 1,2, \ldots , L$. 
Recall that $\scrPZC$ is an orthogonal matrix with $\scrPZC^{-1} =  \scrPZC^T$ and 
denote
\bes  
P(Z,C)= \scrPZC^T\, P\, \scrPZC, \quad P = \scrPZC\, P(Z,C) \, \scrPZC^T.
\ees
According to $Z$ and $C$, matrix $P$ is partitioned into $K^2$ blocks 
$P^{(k_1,k_2)}(Z,C) \in [0,1]^{n_{k_1} \times n_{k_2}}$,
$k_1, k_2 = 1, \ldots, K$,  with the block-averages given by \eqref{eq:B_def}.  
In addition, blocks  $P^{(k_1,k_2)}(Z,C)$ can be combined into the $L^2$ meta-blocks
\bes  
\tilde{P}^{(l_1,l_2)}(Z,C) \in [0,1]^{N_{l_1} \times N_{l_2}},  
\ees
corresponding to probabilities of connections between  meta-communities 
$l_1$ and $l_2$,   $l_1, l_2= 1, \ldots , L$.
\begin{figure} [t]
 \begin{center}
\includegraphics[width = 10cm]{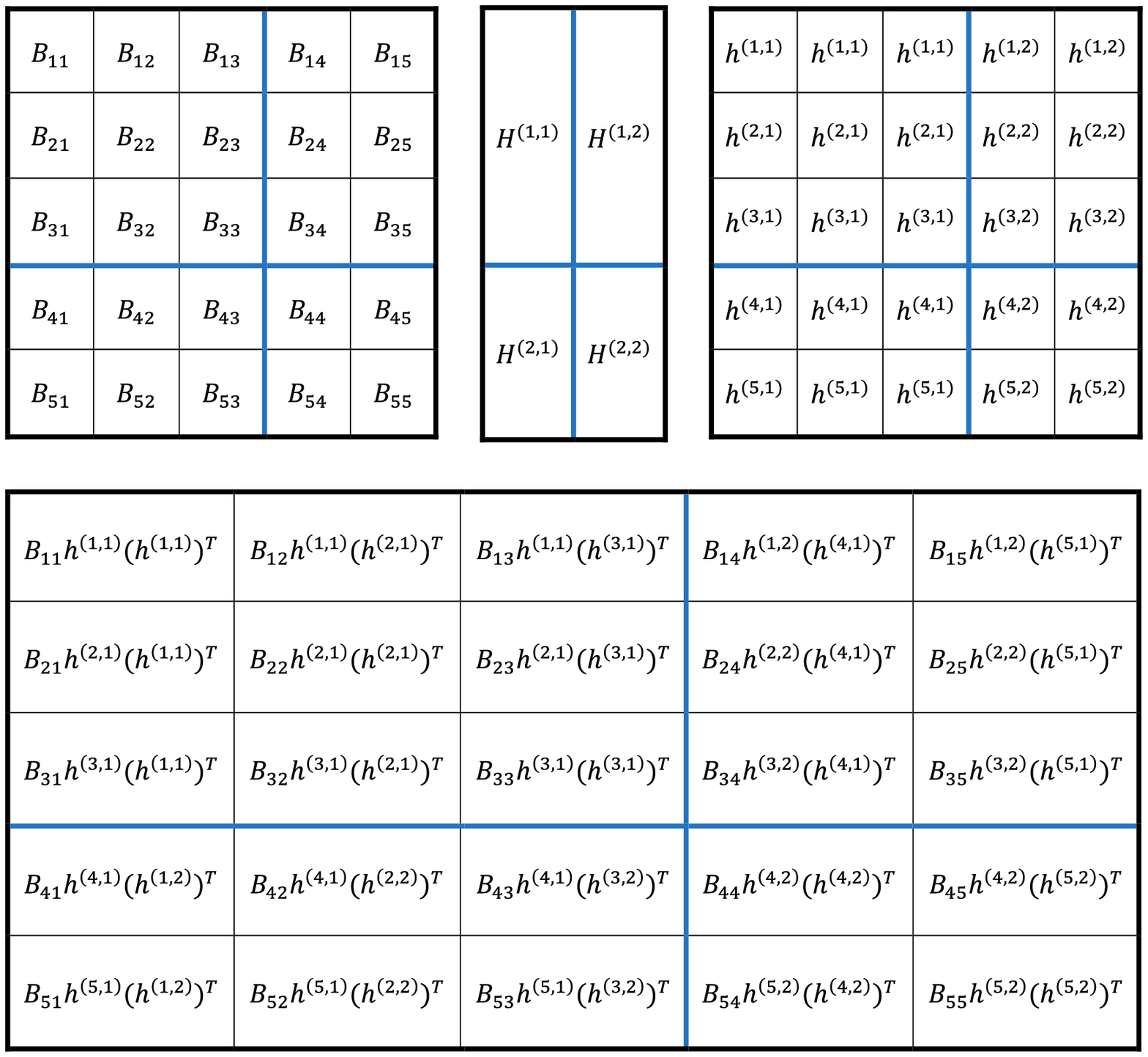} 
\caption{\it {\small Matrices associated with the NBM with $K=5$, $L=2$, $K_1=3$, $K_2=2$.   Bold lines identify the meta-blocks.
Top left: matrix $B$ partitioned into blocks $B^{(l_1,l_2)}$.
Top, middle: matrix $H$. 
Top right: matrix $H$  with columns expressed via vectors $h^{(k,l)}$ and repeated: 
column 1- $K_1$ times; column 2 - $K_2$ times.
Bottom: the probability  matrix  with 
  $K^2$ blocks and $L^2$ meta-blocks.
}}
\label{mn:fig02}
 \end{center}
\end{figure}

Consider matrix   $H \in \RR_+^{n \times L}$   
(Figure~\ref{mn:fig02}, top middle),
where each column $H_l$, $l=1, \ldots, L$, can be 
partitioned into $K$ sub-vectors $h^{(k,l)} \in \RR_{+}^{n_{k}}$ of 
lengths $n_k$, $k= 1, \ldots , K$.
Those sub-vectors are combined into $L$  meta sub-vectors 
${H}^{(m,l)} \in \RR_+^{N_{m}}$ of lengths $N_{m}$, $m= 1, \cdots , L$, 
according to matrix 
$C$, where $N_m$ is defined in  \eqref{eq:num_nodes_in_mega}.
Similarly, matrix $B \in [0,1]^{K \times K}$ of block probabilities is partitioned into sub-matrices 
$B^{(l_1,l_2)} \in [0,1]^{K_{l_1} \times K_{l_2}}$, $l_1, l_2= 1, \cdots , L$.
With these notations, for any $l_1, l_2= 1, \cdots, L$, the $(l_1,l_2)$-th meta-block of $P$ can be presented as 
\be  \label{eq:P_tilde_blocks}
\begin{aligned}
 \tilde{P}^{(l_1,l_2)}(Z,C) = 
 \left( {H}^{(l_1,l_2)}({H}^{(l_2,l_1)})^T \right) \circ  
\left( J^{(l_1)}B^{(l_1,l_2)} (J^{(l_2)})^T \right), 
\end{aligned}
\ee
where $A \circ B$ is the Hadamard product of $A$ and $B$, and matrices 
$J^{(l)} \in \{0,1\}^{N_l \times K_l}$, $l=1, \ldots,L$,  are of the form
\be    \label{eq:Jl_matrices}  
\begin{aligned}
{\small 
 J^{(l)}=
  \begin{bmatrix}
     1_{n_{k_1}} & \boldsymbol{0} & \cdots  & \boldsymbol{0}\\
     \boldsymbol{0} & 1_{n_{k_2}} & \cdots  & \boldsymbol{0}\\
    \vdots & \vdots& \cdots& \vdots\\
    \boldsymbol{0} & \boldsymbol{0} & \cdots  & 1_{n_{k_{K_l}}}\\
  \end{bmatrix}.   
}
\end{aligned}
\ee 
 %

\noindent
By rewriting  \eqref{eq:P_tilde_blocks} in an equivalent form,  one can conclude that each of the
meta-blocks $\tilde{P}^{(l_1,l_2)}(Z,C)$ (and, hence, $\tilde{P}^{(l_1,l_2)}$ if we scramble them 
to the original order) follows the (non-symmetric) DCBM model with $K_{l_1} \times K_{l_2}$ blocks.
Specifically, for a pair of sub-vectors $H^{(l_1,l_2)} \in \RR_+^{N_{l_1}}$ 
and $H^{(l_2,l_1)}\in \RR_+^{N_{l_2}}$ of matrix $H$ 
and a  matrix $B^{(l_1,l_2)} \in [0,1]^{K_{l_1} \times K_{l_2}}$ containing
average probabilities of connections for each pair of communities 
within  the   meta-community $(l_1, l_2)$ one has
\bes 
  \tilde{P}^{(l_1,l_2)}(Z,C) =  
Q^{(l_1,l_2)}  J^{(l_1)}B^{(l_1,l_2)} (J^{(l_2)})^T Q^{(l_2,l_1)}. 
\ees
Here, $Q^{(l_1,l_2)} = \diag (H^{(l_1,l_2)})$ and  the $(k_1,k_2)$-th block of $P(Z,C)$ is given by 
\be \label{eq:P_blocks} 
P^{(k_1,k_2)}(Z,C) = B_{k_1, k_2} h^{(k_1, l_2)} \left(h^{(k_2, l_1)}\right)^T, \quad l_i = c(k_i), \ i=1,2,
\ee 
where   $h^{(k, l)} \in \RR_+^{n_k}$ is a sub-vector of $H^{(m,l)}$ with $m = c(k)$.
Observe that the formulation above imposes a natural scaling on the sub-vectors   
$h^{(k,l)}$  of $H$, since it follows from equations \eqref{eq:B_def}  and \eqref{eq:P_blocks},
that for any pair of communities $(k_1, k_2)$ which belong to a pair of meta-communities $(l_1,l_2)$, one has 
\be \label{eq:B_matr}
 n_{k_1}\, n_{k_2}\ B_{k_1, k_2} =   1_{k_1}^T\, P^{(k_1,k_2)}(Z,C)\,  1_{k_2} = 
 B_{k_1, k_2} \lkr 1_{k_1}^T h^{(k_1, l_2)} \rkr \,  \lkr 1_{k_2}^T\, h^{(k_2, l_1)}\rkr.
\ee 
The latter implies that for any $k=1,\ldots,K$ and $l=1,\ldots,L$,  
\be \label{eq:scaling}
1_{k}^T h^{(k, l)} = n_k, \quad k=1, \ldots, K,\ l=1, \ldots, L.
\ee
\begin{figure} [t]
 \begin{center}
\includegraphics[width = 6cm]{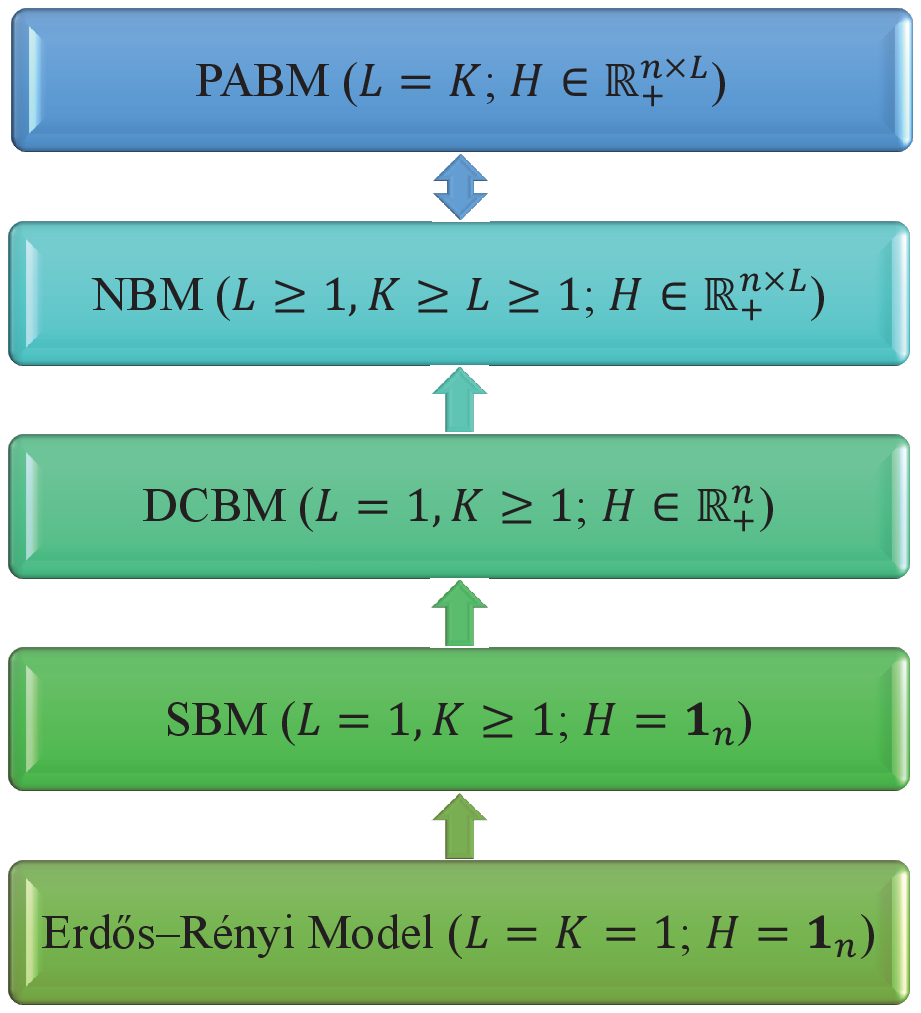} 
\caption{The hierarchy of block models}
\label{mn:fig01}
 \end{center}
\end{figure}
Now, it is easy to see that all block models, the SBM, the DCBM and the PABM, can be
viewed as particular cases of the NBM introduced above. Indeed,  the DCBM is a
particular case of the NBM with $L=1$ while the PABM corresponds to the setting of
$L=K$. Finally, due to \eqref{eq:scaling}, the SBM constitutes a particular case of the NBM 
with $L=1$ and  matrix $H$ reduced to vector $1_n$, the $n$-dimensional 
column  vector  with all entries equal to one. 
Moreover, the absence of the community structure (whether in the SBM or the DCBM) is 
equivalent to $K=1$, and implies that the NBM necessarily reduces to the DCBM. 
This one-community DCBM is indeed just the Chung-Lu model introduced in \cite{Chung15879}. 

\begin{remark} \label{rem:zero_B} {\bf The case of unconnected communities. \ } 
{\rm Note that equations \eqref{eq:P_blocks} and \eqref{eq:B_matr} become identities for any vectors $h^{(k_1, l_2)}$ and 
$h^{(k_2, l_1)}$ if $B_{k_1, k_2} = 0$, 
 $c(k_1) = l_1$ and $c(k_2) = l_2$, which  happens if matrix $P^{(k_1,k_2)}(Z,C)$ is identically equal to zero.
In this case, there are two possibilities. If there exists $\tk_2$ with $c(\tk_2) = l_2$ such that 
$B_{k_1, \tk_2} \neq 0$,  then set $h^{(k_1, l_2)}  =   h^{(k_1, c(\tk_2))}$. 
If no such $\tk_2$ exists (which corresponds to the case when the whole row of matrix $B^{l_1, l_2)}$
is equal to zero), then set  $h^{(k_1, l_2)}  = 1_{n_{k_1}}$. The latter can be interpreted as an
understanding that,  if all nodes in community $k_1$ are not connected to nodes in meta-community $l_2$,
they are ``equally unconnected''. 

Observe that treating zero elements of matrix $B$ in this manner leads to the smallest number of meta-communities and, hence, to 
the smallest number of parameters in the model. For example, in the extreme case  when matrix $B$ is diagonal, one obtains 
that matrix $H$ has only one column, $L=1$ and the NBM just reduces to DCBM. 
%
}
\end{remark}


\section{Optimization procedure for estimation and clustering}
\label{sec:opt_proc}

Note that, in terms of the matrices $J^{(l)}$ defined in \eqref{eq:Jl_matrices}, 
the scaling conditions \eqref{eq:scaling} appear as 
\be \label{eq:alt_scaling}
(J^{(l)})^T \, Q^{(l,l')} \, J^{(l)} = (J^{(l)})^T \,  J^{(l)}, \  l,l'= 1,...,L.
\ee 
Let $\mathscr{P}_{\hZ,\hC}$ be  the permutation matrix corresponding to  
estimated clustering matrices $\hZ \in \calM_{n,\hK}$ and $\hC \in \calM_{\hK,\hL}$. 
Consider the set $\Im (n,K,L)$  of matrices $\Theta$ with blocks $\Theta^{(l_1,l_2)} \in [0,1]^{N_{l_1} \times N_{l_2}}$, $l_1,l_2 = 1,...,L,$  such that
conditions \eqref{eq:num_nodes_in_mega} and  \eqref{eq:alt_scaling} hold and
\begin{align}  
  & \Theta =  \bigcup_{l_1,l_2}   \Theta^{(l_1,l_2)}, \,
    \Theta^{(l_1,l_2)} =  
   Q^{(l_1,l_2)} J^{(l_1)} B^{(l_1, l_2)} (J^{(l_2)})^T  Q^{(l_2,l_1)},  \nonumber \\
 & B^{(l_1, l_2)} \in [0,1]^{K_{l_1} \times K_{l_2}}, \, 
  Q^{(l_1,l_2)} \in \calD_{l_1}, \label{eq:opt_cond} \\
 & Z \in \calM_{n,K}, \  C \in \calM_{K,L}, \, 
l_1,l_2 = 1,...,L,  \nonumber
\end{align}
where $\calD_m$ the set of diagonal matrices with diagonals in $\RR_+^m$.  
 %
 Then, it is easy to see that $P =  \scrPZC^T \, \Theta \, \scrPZC$, so its estimator can be obtained as 
 \be  \label{eq:P_total_est} 
 \hat{P}  =\mathscr{P}_{\hZ,\hC}\  \hThe (\hZ,\hC) \  \mathscr{P}_{\hZ,\hC}^T.
 \ee
Here, for given values of $K$ and $L$,  $(\hZ,\hC, \hThe)$ is a solution of the following optimization problem
\be \label{eq:opt1}
{\small (\hZ,\hC,\hThe) \in  \underset{Z,C,\Theta}{\text{argmin}}  
\left\|A(Z,C)  -  \Theta  \right\|_{F}^2  
}
\ee
subject to conditions $A(Z,C) = \scrPZC^T\, A \, \scrPZC$, 
\eqref{eq:num_nodes_in_mega},  \eqref{eq:alt_scaling} and \eqref{eq:opt_cond}.
In real life, however, the values of $K$ and $L$ are unknown and need to be incorporated into 
the optimization problem by adding  a penalty $\Pen(K,L)$ on $K$ and $L$: 
\be \label{eq:opt2}
\begin{aligned}
 (\hThe, \hZ, \hC,\hat{K}, \hat{L}) \in 
 \underset{Z,C,K,L,\Theta}{\text{argmin}}  
\left\{ 
\left\|A (Z,C)  -  \Theta  \right\|_{F}^2  + \Pen(K,L) \right\},  
\end{aligned}
\ee
where optimization is carried  out subject to conditions  
$A(Z,C) = \scrPZC^T A \scrPZC$,  
\eqref{eq:num_nodes_in_mega},  \eqref{eq:alt_scaling} and \eqref{eq:opt_cond}.
After that, the estimator  $\hat{P}$  of $P_*$ can be obtained as \eqref{eq:P_total_est}.
The penalty in \eqref{eq:opt2} should account for the difficulty of estimating  $n L + K^2$ unknown parameters
($n L$ entries in matrix $H$ and $K^2$  entries in matrix $B$) and uncertainty of clustering 
which is of the logarithmic order $n \ln K$  of the cardinality of the set of clustering matrices. 
For this reason, we choose the penalty   of the form 
\be \label{eq:Pen_Klopp}
\Pen(K,L) = C_1 ( n L + K^2) \ln n + C_2 n \ln K  
\ee
where  $C_1$ and $C_2$ are absolute constants. The logarithmic factor $\ln n$ in 
\eqref{eq:Pen_Klopp} is due to the proof technique and,  can possibly be removed.

In practice, one would need to solve optimization problem \eqref{eq:opt1}
for each $K = 1, ...,n,$ and $L = 1, \ldots, K$, and then find the values $(\hK,\hL)$ that minimize the right 
hand side in \eqref{eq:opt2}. After that, the estimator $\hat{P}$ of $P$ is obtained as \eqref{eq:P_total_est}.
Then, the following statement holds.
%

\begin{theorem}  \label{th:oracle_Klopp}
Let Assumptions A1 and A2 hold.  Let $(\hThe, \hZ, \hC,\hat{K}, \hat{L})$ be a solution of 
optimization problem \eqref{eq:opt2} subject to conditions $A(Z,C) = \scrPZC^T A \scrPZC$, 
\eqref{eq:num_nodes_in_mega},  \eqref{eq:alt_scaling} and \eqref{eq:opt_cond} with  the penalty given by \eqref{eq:Pen_Klopp}.
Then, for the estimator  $\hat{P}$   given by \eqref{eq:P_total_est},   the true matrix $P_*$,  
 any $K$, $L$, $ Z \in \calM_{n,K}$, $C \in \calM_{K,L}$ and any 
 matrix $P = \scrPZC  \Theta \scrPZC^T$ with $\Theta \in \Im (n,K,L)$,  
 one has 
\begin{align*} 
& \PP \lfi  \|\hat{P}-P_{*}\|_F^2   \leq   
3 \big[\|P - P_{*}\|_F^2 + \Pen (K,L)\big ]  \rfi  
  \geq  1 -(n^2 \log_2 n +1) e^{-n/32}, \\
&  \EE\|\hat{P} - P_{*}\|_F^2   \leq     
3 \big[\|P - P_{*}\|_F^2 + \Pen (K,L)\big] 
 + n^5  e^{-n/32}.
\end{align*}
%
\end{theorem}

Solution of optimization problem \eqref{eq:opt2} requires a search over the 
continuum of matrices $\Theta \in \Im (n,K,L)$.
In order to simplify the estimation, we  consider a solution of a more straightforward optimization problem. 
It is easy to observe (see Figure \ref{mn:fig02}) that each of the block 
columns of matrix $P$ is 
a matrix of rank one and, given the clustering, it  can be obtained by the rank one projection  
of the respective adjacency sub-matrix.
Denote the block columns of the re-arranged matrices  $P$ and $A$ by $P^{(l,k)}(Z,C)$  
and $A^{(l,k)}(Z,C)$.
Then, the optimization problem appears as 
\be  \label{eq:opt_main}  
\begin{aligned}
(\hZ, \hC,\hat{K}, \hat{L})  \in   
 \underset{Z,C,K,L}{\text{argmin}} 
\Bigg \{\displaystyle \sum_{l = 1}^L \sum_{k  = 1}^K \left\|A^{(l,k)}(Z,C)  - 
\Pi_{(1)} \lkr A^{(l,k)} (Z,C) \rkr \right\|_{F}^2 +  
\oPen (K,L) \Bigg \} 
 \\
   \text{s.t.}\  A(Z,C) = \scrPZC^T A \scrPZC,\ 
 \end{aligned}
 \ee
where $\Pi_{(1)} \lkr A^{(l,k)} (Z,C) \rkr$  is the rank one projection 
 of the matrix $A^{(l,k)}  (Z,C)$.
Then, $\hThe$ is the block matrix with blocks 
$ \hThe^{(l,k)}= \Pi_{(1)} \lkr A^{(l,k)} (\hZ,\hC) \rkr$,
$l = 1,\cdots,\hat{L}, \ k  = 1,\cdots , \hat{K}$.
Note that the new formulation requires estimation of the larger number of parameters 
 $(n K + K^2)$  versus $(n L + K^2)$ in \eqref{eq:opt2}, so the new penalty  is of the form
\be \label{eq:penalty}
\oPen (K,L) =   \Psi_1 n K + \Psi_2 K^2 \ln n+ \Psi_3 n \ln K,
\ee
where $\Psi_1$, $\Psi_2$, and  $\Psi_3$ are positive absolute constants.

\begin{theorem}  \label{th:oracle}
Let Assumptions A1 and A2 hold. Let $(\hThe, \hZ, \hC,\hat{K}, \hat{L})$ be a solution of optimization problem 
\eqref{eq:opt_main}   with $\oPen (K,L)$   of the  form \eqref{eq:penalty}.
Then, for the estimator  $\hat{P}$  of $P_*$ given by \eqref{eq:P_total_est} 
and any $t >0$,    one has 
\begin{align*} 
& \PP \lfi \left\|\hat{P}  -P_{*}\right\|_F^2  \leq   
\tilde{C}\,   [ \oPen (n,K_{*},L_{*}) +  t]  \rfi   
  \geq  1 - 3 e^{-t},  \\
%
&\EE\left\|\hat{P}  - P_{*}\right\|_F^2  \leq  
\tilde{C}\,  [ \oPen (n,K_{*},L_{*}) + 3]. 
\end{align*} 
Here $K_*$ and $L_*$ are the true number of communities and meta-communities 
and $$\tilde{C} = \tilde{C} (\Psi_1, \Psi_2, \Psi_3) >0$$  
is an absolute constant.  
\end{theorem}

\noindent
Observe that Theorem~\ref{th:oracle} delivers smaller error rates if $K_*/L_* \ll \ln n$, i.e., if $n$ is large.
In addition, for known values of $K$ and $L$, one needs to carry  optimization in \eqref{eq:opt_main}
only over the set of clustering matrices.  In this sense, optimization problem \eqref{eq:opt_main} can be viewed as 
a kind of modularity optimization which has been used for estimation and clustering in the SBM (\cite{Bickel21068}),
the DCBM (\cite{zhao2012consistency}) and the PABM (\cite{RePEc:bla:jorssb:v:80:y:2018:i:2:p:365-386}).   
The deficiency of  this sort of approach is that it is NP-hard and requires some replacement by a computationally
viable method. In our case, this relaxation is provided by a subspace clustering which allows us to find the clustering matrix $C$
and hence detect the meta-communities. Subsequently, we detect the communities within meta-communities using  spectral clustering. 
We describe those procedures in detail in the next section.


\section{Detectability of communities and meta-communities}
\label{sec:detect}

As we have mentioned above, in what follows, we focus on the optimization problem \eqref{eq:opt_main}.  
Observe that the viability of the NBM introduced above relies on the correct detection of communities and meta-communities. 
In order to assess identifiability of clustering matrices $Z$ and $C$, consider a noiseless model where one can observe the 
probability matrix $P_*$ instead  of the adjacency matrix $A$. 
Indeed, if matrices $Z$ and $C$ can be correctly recovered 
(up to permutation of columns),
then matrix $B$ can be obtained by averaging the  probabilities in $P_*(Z,C)$ using formula \eqref{eq:B_matr}.
Furthermore, it follows from \eqref{eq:P_tilde_blocks} that sub-columns $H^{(l_1,l_2)}$   of matrix $H$ can be
obtained by   applying rank one approximations to the Hadamard quotient of $ \tilde{P}^{(l_1,l_2)}(Z,C) $
and $J^{(l_1)}B^{(l_1,l_2)} (J^{(l_2)})^T$.

One however does not need to identify all those quantities in order to estimate clustering matrices $Z$ and $C$.
Optimization problem \eqref{eq:opt_main}  suggests that matrices $Z$ and $C$ can be obtained just on the basis 
of modularity optimization based on the partitions of the adjacency matrix. In order to confirm that the communities and 
meta-communities are detectable, we assume that $K=K_*$ and $L=L_*$ are known and impose the following assumptions:
 \\

\noindent
{\bf A1.}\ \ Matrix $B$ is non-singular with  the smallest  singular value bounded away from zero: $\lambda_{\min}(B) \ge \lambda_0 > 0.$ 
 \\

\noindent 
{\bf A2.}\  \ For each $k=1, \cdots , K$, vectors $h^{(k,l)}$, $l=1, \cdots , L$, are linearly independent.
 \\

\noindent  
Under those assumptions, it is easy to see that the meta-columns 
of matrix $P_*$, corresponding to the $l$-th meta-community, lie in the distinct linear subspace $S_l$ of the dimension 
$K_*$ with the basis defined by $K$ distinct combinations of sub-vectors $h^{(k,l)}$, $k=1,...,K_*$.
For this reason, one can find meta-communities by identifying those subspaces.  
Subsequently, for  finding communities within meta-communities, one notes  that the
$l$-th diagonal  block   of the probability matrix  $\tilde{P}^{(l,l)}(Z_*,C_*)$  in \eqref{eq:P_tilde_blocks},
corresponding to the $l$-th meta-community, $l=1, ..., L$, follows the    DCBM model.
Due to Assumption A2, matrix $B^{(l,l)}$  in \eqref{eq:P_tilde_blocks} is of full rank, which guarantees 
identifiability of communities in the meta-community $l$.  
Specifically, the following statement is true:

 
\begin{lemma} \label{lem:detect}
Let Assumptions {\bf A1}  and   {\bf A2}  hold. Let $K=K_*$ and $L=L_*$  be known.
Let $Z_* \in \calM_{n, K}$ and $C_* \in \calM_{K, L}$ be  the true clustering matrices, while    $Z \in \calM_{n, K}$
and $C  \in \calM_{K, L}$  be  arbitrary clustering matrices. Then, 
\begin{align} 
\sum_{l = 1}^L \sum_{k  = 1}^K  \left\|  P_*^{(l,k)}(Z_*,C_*) \right. & \left. -\  \Pi_{(1)} \lkr P_*^{(l,k)}(Z_*,C_*) \rkr \right\|_{F}^2 
  \label{eq:detect} \\
& \leq \sum_{l = 1}^L \sum_{k  = 1}^K  \left\|P_*^{(l,k)}(Z,C)   -  \Pi_{(1)} \lkr P_*^{(l,k)}(Z,C) \rkr  \right\|_{F}^2 \nonumber
\end{align} 
where, for any matrix $B$, $\Pi_{(1)} (B)$ is its rank one approximation. 
Moreover, equality in \eqref{eq:detect} occurs if and only if matrices  $Z$ and $Z_*$, $C$ and $C_*$ coincide up to a permutation of columns.
\end{lemma}

Lemma~\ref{lem:detect} ensures that, if the optimization problem \eqref{eq:opt_main} is applied to the true probability matrix 
with known $K$ and $L$, then the true clustering matrices $Z_*$ and $C_*$ will be recovered up to the permutation of columns.
However, since optimization procedures in \eqref{eq:opt2} and  \eqref{eq:opt_main}  are NP-hard, they cannot be implemented in practice.


\section { Implementation of clustering }
\label{sec:clustering}

In this section, we describe a computationally tractable  clustering procedure that 
can replace optimization procedures in \eqref{eq:opt2} and  \eqref{eq:opt_main}.
Since the model requires  identification  of meta-communities and regular communities,
naturally, the clustering is carried out in two steps. 
First, we find the clustering matrix $C$ that arranges the nodes into $L$ meta-communities. 
Subsequently, we detect communities within  each of the  meta-communities, obtaining the clustering matrix $Z$.

In order to accomplish  the first task, we   observe that, under Assumptions A1 and  A2, the meta-columns 
of matrix $P$, corresponding to the $l$-th meta-community, lie in the distinct linear subspace $S_l$ of the dimension 
$K$ with the basis defined by $K$ distinct combinations of subvectors $h^{(k,l)}$, $k=1,...,K$. 
For this reason, one can find meta-communities by identifying those subspaces. This can be done by 
subspace clustering, the technique which has been well developed by the computer vision community.
Subsequently, for  finding communities within meta-communities, one notes  that the
probability matrix of each meta-community follows the non-symmetric  DCBM model,
 for which there exist several clustering methods.

Subspace clustering is designed for separation of points that lie in the union of subspaces. 
Let $\{ X_{j} \in \R^{D} \}_{j=1}^{n} $ be a given set of points drawn from an unknown union of 
$K \geqslant 1$ linear or affine subspaces $\{ S_{i} \}_{i=1}^{K} $ of unknown dimensions 
$d_{i}= \text{dim}(S_{i})$, $0<d_{i} <D$, $i=1,...,K$. In the case of linear subspaces, the subspaces can be described as 
$S_{i}=\{ \bm{x} \in \R^{D} : \bm{x}=  \bm{U}_{i}\bm{y} \}$, $i=1,...,K,$ 
where $\bm{U}_{i} \in \R^{D \times d_{i}}$ is a basis for subspace $S_{i}$ 
and $\bm{y} \in \R^{d_{i}}$ is a low-dimensional representation for point $\bm{x}$. 
The goal of subspace clustering is to find the number of subspaces $K$, their dimensions 
$\{ d_{i} \}_{i=1}^{K}$, the subspace bases $\{ \bm{U}_{i} \}_{i=1}^{K}$, and the segmentation 
of the points according to the subspaces.

Several methods have been developed to implement subspace clustering such as  algebraic methods 
(\cite{inproceedings}, \cite{Ma:2008:ESA:1405158.1405160}, \cite{vidal2005generalized}),
iterative methods (\cite{P.AgarwalandN.Mustafa2004},
\cite{Bradley:2000:KC:596077.596262},  \cite{tseng2000nearest}), 
and self representation  based methods (\cite{Vidal:2009aa}, 
\cite{Elhamifar:2013:SSC:2554063.2554078}, 
\cite{Favaro:2011:CFS:2191740.2191857},  \cite{Liu:2013:RRS:2412386.2412936}, 
\cite{Liu2010RobustSS},  \cite{soltanolkotabi2014},  
 \cite{vidal2011subspace}).

 %
%
%
\ignore{
Let $\{ X_{j} \in \R^{D} \}_{j=1}^{n} $ be a given set of points drawn from an unknown union of 
$K \geqslant 1$ linear or affine subspaces $\{ S_{i} \}_{i=1}^{K} $ of unknown dimensions 
$d_{i}= \text{dim}(S_{i})$, $0<d_{i} <D$, $i=1,...,K$. In the case of linear subspaces, the subspaces can be described as 
$S_{i}=\{ \bm{x} \in \R^{D} : \bm{x}=  \bm{U}_{i}\bm{y} \},$ $i=1,...,K,$
where $\bm{U}_{i} \in \R^{D \times d_{i}}$ is a basis for subspace $S_{i}$ 
and $\bm{y} \in \R^{d_{i}}$ is a low-dimensional representation for point $\bm{x}$. 
The goal of subspace clustering is to find the number of subspaces $K$, their dimensions 
$\{ d_{i} \}_{i=1}^{K}$, the subspace bases $\{ \bm{U}_{i} \}_{i=1}^{K}$, and the segmentation 
of the points according to the subspaces. 
}
%
%
In this paper we  use  the self-representation type method, the Sparse Subspace
Clustering (SSC) developed by  \cite{Elhamifar:2013:SSC:2554063.2554078}. The technique
is based on   representation of each of the vectors as a sparse linear combination of
all other vectors, with the expectation that a vector is more likely to be represented
as a linear combination of vectors in its own subspace rather than other subspaces.   
The weights obtained by this procedure are used to form the affinity matrix which, in turn, is partitioned using the spectral clustering methods.

If matrix $P_*$ were known, the weight matrix $W$ would be based on writing every data point 
as a sparse linear combination of all other points by minimizing the number of nonzero coefficients 
\begin{equation}  \label{mn:opt_prob1}
\mathop\text{min}_{W_{j}}  \left\|W_{j}\right\|_{0}    \quad   \mbox{s.t.}    \quad   (P_*)_{j}=\sum_{k \ne j} W_{k,j} (P_*)_{k}
\end{equation}
where, for any matrix $B$,  $B_{j}$ is its   $j$-th column. The affinity matrix  
of the SSC is the symmetrized version of the weight matrix $W$. 
Note that since, due to Assumption {\bf  A2},  the subspaces are linearly independent, 
the solution to the optimization problem  \eqref{mn:opt_prob1}  
is $W_j$ such that $W_{k,j}\neq 0$ only if points $k$ and $j$ are in the same subspace. 
Since the problem \eqref{mn:opt_prob1}  is NP-hard, one usually solves its convex
LASSO relaxation
\begin{equation}  \label{mn:opt_prob2}
\mathop\text{min}_{W_{j}}  \left\|W_{j}\right\|_{1}  \quad   \mbox{s.t.}  \quad  (P_*)_{j}=\sum_{k \ne j} W_{k,j} (P_*)_{k}
\end{equation}
In the case of data contaminated by noise, the SSC algorithm does not attempt to write  
data as an exact linear combination of other points and replaces   \eqref{mn:opt_prob2}
by penalized optimization. 
Specifically, in our simulations, we solve  the elastic net problem
\begin{equation} \label{mn:Elastic_Net}
\begin{aligned}
 \widehat{W}_j  \in  \underset{W_{j}}{\text{argmin}}   
\lfi \Big [ 0.5 \left\|{A_{j}-AW_{j}}\right\|_{2}^{2} + \gamma_1 \left\|W_{j}\right\|_{1} \right. 
 \left. + \gamma_2 \left\|W_{j}\right\|_{2}^{2} \Big ] 
\   \mbox{s.t.}  \  W_{j,j}=0  \rfi, 
 \ j=1,\ldots,n, \\
\end{aligned}
\end{equation}
where $\gamma_1, \gamma_2 > 0$ are tuning parameters. The quadratic term stabilizes the LASSO problem by 
making the problem strongly convex. 
We solve  \eqref{mn:Elastic_Net} using a fast version of the LARS algorithm 
implemented  in  SPAMS Matlab toolbox  \cite{mairal2014spams}.  
Given $\widehat{W}$, the  clustering matrix  $C$ is then obtained by applying spectral clustering 
to the affinity matrix   $|\widehat{W}| + |\widehat{W}^{T}|$, 
where, for any matrix $B$, matrix $|B|$ has absolute values of elements of $B$ as its entries.
Algorithm \ref{mn:SSC:alg} summarizes the SSC procedure described above.

The correctness of the SSC relies on the   so called {\it self-expressiveness property}  (SEP), 
which guarantees that each column of the probability matrix $P_*$ will be represented using columns of its own subspace rather 
than columns of the other subspaces. 
The latter leads to the $n \times n$ estimated matrix of weights $\widehat{W}$ where $\widehat{W}_{i,j} =0$ if nodes $i$ and $j$ 
are in different meta-communities. Subsequently, according to Algorithm~1, one applies spectral clustering to the symmetrized 
matrix of weights $|\widehat{W}| + |\widehat{W}^T|$. It is easy to see that, if the true matrix of probabilities  $P_*$  
were available, then, under Assumptions A1 and A2, matrix $\widehat{W}$ obtained as a solution of \eqref{mn:opt_prob1} 
or  \eqref{mn:opt_prob2},   satisfies the SEP.  Since matrix $A$ is generated on the basis of  matrix  $P_*$,
one expects that the entries of matrix  $\widehat{W}$, obtained as a solution of \eqref{mn:Elastic_Net}, are equal 
to zero for pairs of nodes that belong to different meta-communities. Although the latter fact is supported by simulations, the 
formal proof of this statement is very nontrivial and is not presented in this paper. 
%
%
%
\begin{algorithm} [t] 
\caption{\ The SSC procedure}
\label{mn:SSC:alg}
\begin{flushleft} 
{\bf Input:} Adjacency matrix $A$, number of meta-communities $L$, tuning parameters $\gamma_1,\gamma_2$\\
{\bf Output:} Clustering matrix $C$ \\
{\bf Steps:}\\
{\bf 1:} For $j=1, ...,n$, find $\widehat{W}_j$  in \eqref{mn:Elastic_Net}\\
{\bf 2:}  Apply   spectral clustering to the affinity matrix $|\widehat{W}| + |\widehat{W}^{T}|$ to find clustering matrix   $C$ 
\end{flushleft} 
\end{algorithm}
%
%
 
Once the meta-communities are discovered, one needs  to detect communities inside of each meta-community. 
Recall that each meta-community  follows the non-symmetric DCBM.
One of the popular clustering methods for the  DCBM is the weighted $k$-median algorithm 
used in  \cite{lei2015} and \cite{gao2018community}. 
Algorithm \ref{mn:SC_k_median:alg} follows \cite{gao2018community}.
For the known number of communities $K$,  the algorithm starts with 
estimating the probability matrix $P$ by the best rank $K$ approximation of the adjacency matrix, obtaining  $\hat{P}=UDU^T$, where $U \in \RR^{n \times K}$ 
contains $K$ leading eigenvectors and $D$ is a diagonal matrix of top $K$ eigenvalues. 
After that, the columns of  $\hat{P}$ are normalized, leading to
$\tilde{P}_i=\hat{P}_i/\|\hat{P}_i\|_1$, $i=1,2,\ldots,n$. 
Finally, the  $k$-median  algorithm is applied to $\tilde{P}$ 
to find the community assignment. 
%
%
\begin{algorithm} 
\caption{\ Spectral clustering with $k$-median}
\label{mn:SC_k_median:alg}
\begin{flushleft} 
{\bf Input:} Adjacency matrix $A \in \{0,1\}^{n\times n}$, number of clusters $k$\\
{\bf Output:} Community assignment \\
{\bf Steps:}\\
{\bf 1:} Find $\hat{P} = U D U^T$ with $U \in \RR^{n \times K}$,   the best rank $k$ approximation of matrix $A$ \\
{\bf 2:} For $j=1, ...,n$, find  $\tilde{P}_j=\hat{P}_j/\|\hat{P}_j\|_1$\\
{\bf 3:}  Apply   $k$-median  algorithm to $\tilde{P}$ to obtain community assignment
\end{flushleft} 
\end{algorithm}
%
%
%

In the first step of clustering, we apply Algorithm \ref{mn:SSC:alg} 
to the adjacency matrix $A$ 
to find $L$ meta-communities defined by the clustering matrix $C$.
In the second step, Algorithm \ref{mn:SC_k_median:alg} is  applied to each of $L$ meta-communities, 
obtained at the first step. Specifically, we apply Algorithm \ref{mn:SC_k_median:alg} with $k=K_l$ and $n=N_l$ 
to cluster the $l$-th meta-community,   $l=1,...,L$.
The union of these communities combined with the clustering matrix $C$,
yields the clustering matrix $Z$.
 We elaborate on the implementation of this two-step clustering procedure in Section~\ref{sec:simulations}.

\begin{remark}
\label{rem:findKL} {\bf Finding the number of communities and meta-communities. \ } 
{\rm In theory, in order to find the unknown  values of $K$ and $L$, one needs to  solve optimization problem 
\eqref{eq:opt2} or \eqref{eq:opt_main}  
for each $K = 1, ...,n$ and $L = 1, \ldots, K$, and then find the values
$(\hK,\hL)$ that minimize the right hand side in \eqref{eq:opt2} or
\eqref{eq:opt_main}.  In practice, however, the constants in the
penalties are too large and will lead to significant underestimation of
the number of communities and meta-communities. For this reason, in
practice, one should run optimization with several small values of $L$
(say, $L=1,2,3$). For each of the values of $L$, one finds the meta-communities using SSC (Algorithm~1). As soon as   the
meta-communities are identified, each of those meta-communities follow
the DCBM,  hence, the problem reduces to finding the number of communities in those DCBMs. Several authors tackled this problem, see,
e.g., \cite{ma2019determining}. Subsequently, one can choose the number
of meta-communities using a common complexity penalty such as AIC or BIC.
}
\end{remark}

\begin{remark}
\label{rem:alt_clust} {\bf Alternative way of clustering. \ }  
{\rm
Under the assumptions of the paper, if the SSC was applied to the matrix $P_*$
instead of the adjacency matrix $A$, it  would yield  $W_{i,j}=0$  when nodes $i$ and $j$ are in the 
same meta-community but different communities. Hence, it is possible to reverse the procedure and first 
cluster nodes into the communities using the SSC and then, subsequently cluster the communities into the meta-communities.
However, since   clustering is carried out on the basis of matrix $A$, and the meta-communities are in general larger than communities
(and there are fewer of them), the procedure used in the paper is more precise and stable, so that the estimated weights are more likely to 
satisfy the self-expressiveness property (SEP). 
}
\end{remark}


\section{Simulations and a real data example}
\label{sec:simulations_real_data} 


\subsection{Simulations on synthetic networks}
\label{sec:simulations}

In the experiments with synthetic data, we generate networks with $n$ nodes, 
$L$ meta-communities and  $K$ communities  that fit the NBM. 
For simplicity, we consider perfectly balanced networks where the number of nodes in each community and 
meta-community are respectively  $n/K$ and  $n/L$, and there are  $K/L$ communities in each meta-community. 
First, we  generate $L$ distinct $n$-dimensional random vectors with entries 
between 0 and 1. 
%
%
To this end, we generate a random vector $Y \in (0,1)^n$
and partition it into $K$ blocks $Y^{(k)}$, $k=1,...,K$, of   size $n/K$. The vector $\bar{h}^{(1)}$ 
is generated from $Y$ by sorting each block of $Y$ in ascending order.
After that, we partition each of the $K$ blocks,  $\bar{h}^{(k,1)}$  of $\bar{h}^{(1)}$,
into $L$ sub-blocks $\bar{h}^{(k,1)}_i$,  $i=1,...,L$, of equal size. 
To generate the $k$-th block $\bar{h}^{(k,2)}$ of $\bar{h}^{(2)}$, we reverse the order 
of entries in each sub-block  $\bar{h}^{(k,1)}_i$ and rearrange them in descending order.
The blocks $\bar{h}^{(k,s)}$ of subsequent vectors $\bar{h}^{(s)}$, $s=3, ...,L$, are 
formed by re-arranging the order of sub-blocks $\bar{h}^{(k,2)}_i$ in each sub-vector 
$\bar{h}^{(k,2)}$.
The $L$ vectors $\bar{h}^{(l)}$, $l=1,...,L$, generated by this procedure have different 
patterns leading to detectable meta-communities. Subsequently, we scale the vectors as  
$H^{(k,l)}=
(n/K)\, \bar{h}^{(k,l)}/\|\bar{h}^{(k,l)}\|_1$, 
$k=1,...,K$,    $l=1,...,L$,  obtaining matrix $H$.
After that,  we replicate $K/L$ times each  of the 
columns of $H$   (Figure \ref{mn:fig02}, top right) 
and denote the resulting matrix by  $\tilde{H}$.
Matrix $B$ has entries  
\be  \label{mn:matrix_B}  
B_{k,l}=\tilde{B}_{k,l}  \lkv(\tH_{max})_{k,l}\rkv^{-2},  \quad  k,l=1,...,K,
\ee 
where $\tilde{B}$ is a $(K \times K)$ symmetric matrix with random 
entries between 0.35 and 1 to avoid very sparse networks, 
and the largest entries of each row (column) are on the diagonal. Matrix  $\tH_{max}$ is a 
$K \times K$ symmetric matrix defined as
\bes 
(\tH_{max})_{k,l}= \max \left (\tH^{(k,l)},\tH^{(l,k)} \right ),  \quad  k,l=1,...,K,
\ees
where $\tH^{(k,l)}$ is the $(k,l)$-th block of matrix $\tH$. 
The term $\lkv(\tH_{max})_{k,l}\rkv^{-2}$ in \eqref{mn:matrix_B} 
guarantees that the entries of probability matrix $P(Z,C)$ do not exceed one.  
To control   how assortative the network is, we multiply the off-diagonal entries of $B$ 
by the parameter $\omega \in (0,1)$. 
The values of $\omega$ close to zero produce an almost block diagonal probability 
matrix $P(Z,C)$ while the values of $\omega$ close to one 
lead to $P(Z,C)$ with more diverse entries. We  obtain the probability matrix 
$P(Z,C)$ as
\bes 
P^{(k,l)}(Z,C)=B_{k,l}\, \tH^{(k,l)} \left ( \tH^{(l,k)} \right ) ^T ,  \   k,l=1,...,K.
\ees
After that, to obtain the probability matrix $P$, we   generate 
random clustering matrices $Z \in \calM_{n,K}$ and $C \in \calM_{K,L}$ and 
their corresponding $n \times n$ permutation matrices 
$\scrP(Z)$ and $\scrP(C)$, respectively. Subsequently, we set 
$\scrPZC=\scrP(Z)\scrP(C)$ and obtain the  
probability matrix $P$ as $P=\scrPZC P(Z,C) (\scrPZC)^T$. 
Finally we generate the lower half of the adjacency matrix $A$ as independent 
Bernoulli variables  
$A_{i,j} \sim \text{Bern}(P_{i,j})$, $i=1, \ldots, n, j=1, \ldots, i-1$, 
and set  $A_{i,j} = A_{j,i}$ when $j >i$. 
In practice, the diagonal $\text{diag}(A)$ of matrix $A$ is unavailable, 
so we estimate matrix $P$ without its knowledge.

We apply Algorithm~\ref{mn:SSC:alg} to find the clustering matrix $\hC$. 
Since the diagonal elements of matrix $A$ are unavailable, we initially set $A_{i,i}=0$, $i=1,...,n$. 
We use $\gamma_1=30\rho(A)$ and $\gamma_2=125(1-\rho(A))$ where $\rho(A)$ is the density 
of matrix $A$, the proportion of nonzero entries in $A$.
The spectral clustering in step 2 of the Algorithm~\ref{mn:SSC:alg} is carried out by 
the normalized cut algorithm of \cite{shi2000normalized}. 
Once  the meta-communities are obtained,  we  apply Algorithm~\ref{mn:SC_k_median:alg} 
to detect communities inside each meta-community. The union of detected communities    
and the clustering matrix $\hC$ yields the clustering matrix $\hZ$.
Given $\hZ$ and $\hC$, we generate matrix 
$A(\hZ,\hC) = \scrP_{\hZ,\hC}^T\, A \, \scrP_{\hZ,\hC}$ 
with blocks $A^{(k,l)}( \hZ,\hC )$,
$k =1,\ldots, K$, $l=1, \ldots, L$, and obtain $\widehat{\Theta}^{(k,l)}(\hZ, \hC)$ 
by using the rank one projection for each of the blocks. 
Finally, we  estimate  matrix $P$ by $\hat{P}$, given by  formula~\eqref{eq:P_total_est}.



\begin{table} 
\label{mn:table1}
\begin{center}
\begin{tabular}{|c| c|c |c| c|  }
\multicolumn{5}{ l }{  }\\
\hline \hline 
 & \multicolumn{4}{ |c |}{\textbf{DCBM: $K=6$ , $L=1$}} \\
\hline
 & \multicolumn{2}{ |c |}{$n=900$} & \multicolumn{2}{c|}{$n=1260$}\\
\hline
   &  $\omega=0.6$   & $\omega=0.8$    &   $\omega=0.6$   & $\omega=0.8$   \\
\cline{1-5} 
 PABM: Clust. Err.     & 0.82411 (0.00221)   & 0.82378 (0.00210)   & 0.82696 (0.00176)   & 0.82725 (0.00179)      \\  \cline{1-5}
 \textbf{DCBM: Clust. Err.}     & \textbf{0.07326 (0.01389)}  & \textbf{0.12241 (0.03542)}   &  \textbf{0.05302 (0.01040)}  & \textbf{0.08989 (0.02555)}       \\   \cline{1-5}     
 NBM: Clust. Err.    &  0.07326 (0.01389)  &  0.12241 (0.03542)  & 0.05302 (0.01040)   & 0.08989 (0.02555)     \\   \cline{1-5}
 PABM: Est.  Err.    & 0.00885 (0.00089)   & 0.00761 (0.00065)   &  0.00808 (0.00076)  &  0.00693 (0.00078)      \\     \cline{1-5}       
 \textbf{DCBM: Est.  Err.}     & \textbf{0.00212 (0.00016)}  & \textbf{0.00274 (0.00023)}   &  \textbf{0.00153 (0.00010)}  & \textbf{0.00196 (0.00015)}    \\      
\cline{1-5}    
NBM: Est.  Err.     & 0.00203 (0.00010)  & 0.00237 (0.00014)   &  0.00146 (0.00006)  & 0.00167 (0.00009)       \\      
\cline{1-5}    
\hline
\hline
%
\hline 
 & \multicolumn{4}{ |c |}{\textbf{NBM: $K=6$ , $L=2$}} \\
\hline
 & \multicolumn{2}{ |c |}{$n=900$} & \multicolumn{2}{c|}{$n=1260$}\\
\hline
   &  $\omega=0.6$   & $\omega=0.8$    &   $\omega=0.6$   & $\omega=0.8$   \\
\cline{1-5} 
 PABM: Clust. Err.     & 0.43885 (0.03879)   & 0.45456 (0.02452)   & 0.42849 (0.05346)   & 0.47122 (0.04774)      \\  \cline{1-5}
 DCBM: Clust. Err.    & 0.38644 (0.02658)   & 0.46126 (0.05912)   & 0.39413 (0.01859)   & 0.45976 (0.05440)       \\   
 
 \cline{1-5}  
  
\textbf{ NBM: Clust. Err.}  & \textbf{0.04737 (0.01193)}  & \textbf{0.09207 (0.02935)}   &  \textbf{0.03458 (0.01208)}  & \textbf{0.07601 (0.03838)}    
   \\ \textbf{Communities}  & & & &
   \\   \cline{1-5}
   
NBM: Clust. Err.   &  0.00026 (0.00086)   &  0.00004 (0.00020)  & 0.00458 (0.00759)   & 0.00071 (0.00177)  
\\ Meta-Communities  & & & &
   \\   \cline{1-5}
 
 PABM: Est.  Err.    & 0.00487 (0.00040)   & 0.00507 (0.00033)   &  0.00397 (0.00049)  &  0.00390 (0.00048)      \\     \cline{1-5}       
 DCBM: Est.  Err.     & 0.00518 (0.00084)  & 0.00788 (0.00146)   &  0.00492 (0.00055)  & 0.00733 (0.00101)    \\      
\cline{1-5}    
\textbf{NBM: Est.  Err.}     & \textbf{0.00237 (0.00012)}  & \textbf{0.00284 (0.00019)}   &  \textbf{0.00182 (0.00027)}  & \textbf{0.00201 (0.00011)}        \\      
\cline{1-5}    
\hline
%
\hline 
\hline 
 & \multicolumn{4}{ |c |}{\textbf{NBM: $K=6$ , $L=3$}} \\
\hline
 & \multicolumn{2}{ |c |}{$n=900$} & \multicolumn{2}{c|}{$n=1260$}\\
\hline
   &  $\omega=0.6$   & $\omega=0.8$    &   $\omega=0.6$   & $\omega=0.8$   \\
\cline{1-5} 
 PABM: Clust. Err.     & 0.36722 (0.07693)   & 0.46137 (0.04546)   & 0.40421 (0.05570)   & 0.45825 (0.04972)      \\  \cline{1-5}
 DCBM: Clust. Err.    & 0.28004 (0.06783)   & 0.45044 (0.10014)   & 0.27603 (0.06227)   & 0.44704 (0.08994)       \\   
 
 \cline{1-5}  
  
\textbf{ NBM: Clust. Err.}  & \textbf{0.07400 (0.03607)}  & \textbf{0.06941 (0.04866)}   &  \textbf{0.08807 (0.03832)}  & \textbf{0.09074 (0.07109)}    
   \\ \textbf{Communities}  & & & &
   \\   \cline{1-5}
   
NBM: Clust. Err.   &  0.05104 (0.03903)   &  0.00856 (0.02127)  & 0.07598 (0.04009)   & 0.02564 (0.03323)  
\\ Meta-Communities  & & & &
   \\   \cline{1-5}
 
 PABM: Est.  Err.    & 0.00536 (0.00074)   & 0.00637 (0.00090)   &  0.00452 (0.00053)  &  0.00507 (0.00114)      \\     \cline{1-5}       
 DCBM: Est.  Err.     & 0.00586 (0.00073)  & 0.00936 (0.00125)   &  0.00501 (0.00069)  & 0.00826 (0.00118)    \\      
\cline{1-5}    
\textbf{NBM: Est.  Err.}     & \textbf{0.00363 (0.00072)}  & \textbf{0.00338 (0.00074)}   &  \textbf{0.00318 (0.00070)}  & \textbf{0.00288 (0.00078)}        \\      
\cline{1-5}    
\hline
\hline
%
 \hline 
 & \multicolumn{4}{ |c |}{\textbf{PABM: $K=6$ , $L=6$}} \\
\hline
 & \multicolumn{2}{ |c |}{$n=900$} & \multicolumn{2}{c|}{$n=1260$}\\
\hline
   &  $\omega=0.6$   & $\omega=0.8$    &   $\omega=0.6$   & $\omega=0.8$   \\
\cline{1-5} 
 \textbf{PABM: Clust. Err.}     & \textbf{0.08059 (0.02294)}   & \textbf{0.03141 (0.02969)}   & \textbf{0.05667 (0.02857)}   & \textbf{0.02376 (0.04287)}      \\  \cline{1-5}
 DCBM: Clust. Err.     & 0.20037 (0.01484)  & 0.22489 (0.01141)   &  0.19725 (0.01416)  & 0.21728 (0.01119)       \\   \cline{1-5}     
 NBM: Clust. Err.    &  0.08059 (0.02294)  &  0.03141 (0.02969)  & 0.05667 (0.02857)   & 0.02376 (0.04287)     \\   \cline{1-5}
 \textbf{PABM: Est.  Err.}    & \textbf{0.00434 (0.00021)}   & \textbf{0.00463 (0.00107)}   &  \textbf{0.00308 (0.00057)}  &  \textbf{0.00325 (0.00099)}      \\     \cline{1-5}       
 DCBM: Est.  Err.     & 0.00475 (0.00074)  & 0.00790 (0.00089)   &  0.00438 (0.00051)  & 0.00732 (0.00090)    \\      
\cline{1-5}    
NBM: Est.  Err.     & 0.00433 (0.00020)  & 0.00463 (0.00107)   &  0.00307 (0.00057)  & 0.00325 (0.00099)       \\      
\cline{1-5}    
\hline
\hline
\end{tabular} 
\end{center}
\caption{The clustering errors $\Err (\hZ,Z)$, $\Err (\hC,C)$, and estimation error $\Delta(\hat{P})$ given by~\eqref{eq:AIC},
 for model fitting using the DCBM, the NBM and the PABM,
under correctly and incorrectly specified models. The networks are generated under three different models: 
the DCBM (with $K=6$, $L=1$), the NBM (with $K=6$, $L=2$ and  3), and the PABM (with $K=6$, $L=6$), 
for $n=900$ and 1260 and $\omega=0.6$ and 0.8. The results are evaluated over 30 simulation runs. 
The corresponding standard deviations are given in parentheses.
}
\end{table}


For evaluation of the performance of our method, we generate networks from three different models: 
the DCBM (with $K=6$, $L=1$), the NBM (with $K=6$, $L=2$ and  3), and the PABM (with $K=6$, $L=6$), 
for $n=900$ and $n=1260$ and $\omega=0.6$ and $\omega=0.8$. Then we fit the DCBM, the NBM, and the PABM 
to each of the  generated networks. The proportion of misclassified nodes (clustering error) is  evaluated as 
 \be \label{eq:misclustered}
\Err(Z, \hZ) =  (2n)^{-1}\, \underset{\mathscr{P}_K \in \mathcal{P}_K} {\min}  \| Z \mathscr{P}_K - \hZ \| _F^2  
 \ee 
where $\mathcal{P}_K$ is the set of permutation matrices  
$\mathscr{P}_K: \{1,...,K\} \to  \{1,...,K\}$.
Table \ref{mn:table1} shows the accuracy of clustering for fitting correct  and incorrect  models to the generated networks. 
For all settings, the clustering errors of fitting the correct model  are smaller than those of the incorrect ones. 
Moreover, in the NBM, since meta-communities are detected first, the accuracy of detecting $K$ communities 
depends  on the precision of detecting $L$ meta-communities. One can see from Table \ref{mn:table1} that, when the NBM is the true model, 
there is a significant improvement in the accuracy of detecting $K$ communities using the two-step clustering procedure,
 with finding the meta-communities being the key task. 
It is also worth noting that when DCBM is the true model of the networks, then there is only one
meta-community. Hence, when NBM is fitted to the networks, there is no need to detect meta-communities
(as there is only one). Hence,  we just   detect
communities by applying Algorithm 2, which leads to the   results identical to  the ones obtained by
fitting the true model (DCBM). Similarly, when the true model of the network is PABM, there is only one community inside of each meta-community. Thus, when
NBM is fitted to the networks, 
one only needs to detect meta-communities using Algorithm 1, attaining the same results 
as the ones obtained by fitting the true model (PABM).

Since the model with larger number of parameters allows for a  more accurate estimation of matrix $P$, 
we measure the accuracy of an estimator $\hat{P}$ of   $P$ by the squared Frobenius norm of their  difference 
with the added AIC-type  penalty
 \be \label{eq:AIC}
\Delta(\hat{P}) =  n^{-2}\left\{ \|  \hat{P} - P \| _F^2 + 2 \bar{P}\ N_{Par} \right\}.
 \ee 
Here, $ \|{\hat{P}-P}\|_{F}^{2}$ acts as a pseudo likelihood, $\bar{P}=n^{-2} \sum_{i,j = 1}^n P_{ij}$ is
the average density of $P$,  and $N_{Par}$ is the number of parameters in a model:
 $N_{Par}=n+K(K+1)/2-1$ for the DCBM, $N_{Par}=nL+K(K+1)/2-KL$ for the NBM, and $N_{Par}=nK$ for the PABM. 
In DCBM, $\hat{P}$ is obtained by solving a low rank approximation problem, as it is  explained in \cite{gao2018community}.
In the PABM, $\hat{P}$ is found by the post-clustering estimation, which is based on rank one approximations 
(see \cite{noroozi2019sparse}). 
Table \ref{mn:table1} shows that, even if the AIC penalty on the number of parameters is added, 
  correctly fitted models have smaller estimation error \eqref{eq:AIC} than incorrectly fitted ones.

Thus, the results in Table \ref{mn:table1} can be summarized as follows. 
If the true model is NBM, then NBM fits best and the other models fit rather poorly. On the other hand,
if the true model is DCBM (or PABM), then DCBM (or PABM) fits best, but NBM also fits well, with
accuracy not much worse than the true model. Therefore, without knowledge of the true model
(as it happens in real-world scenarios), fitting NBM is the safest option.


\subsection{Real data examples}
\label{sec:real_data} 

\begin{figure}[t]
   \begin{center}
\[ \includegraphics[height=7cm] {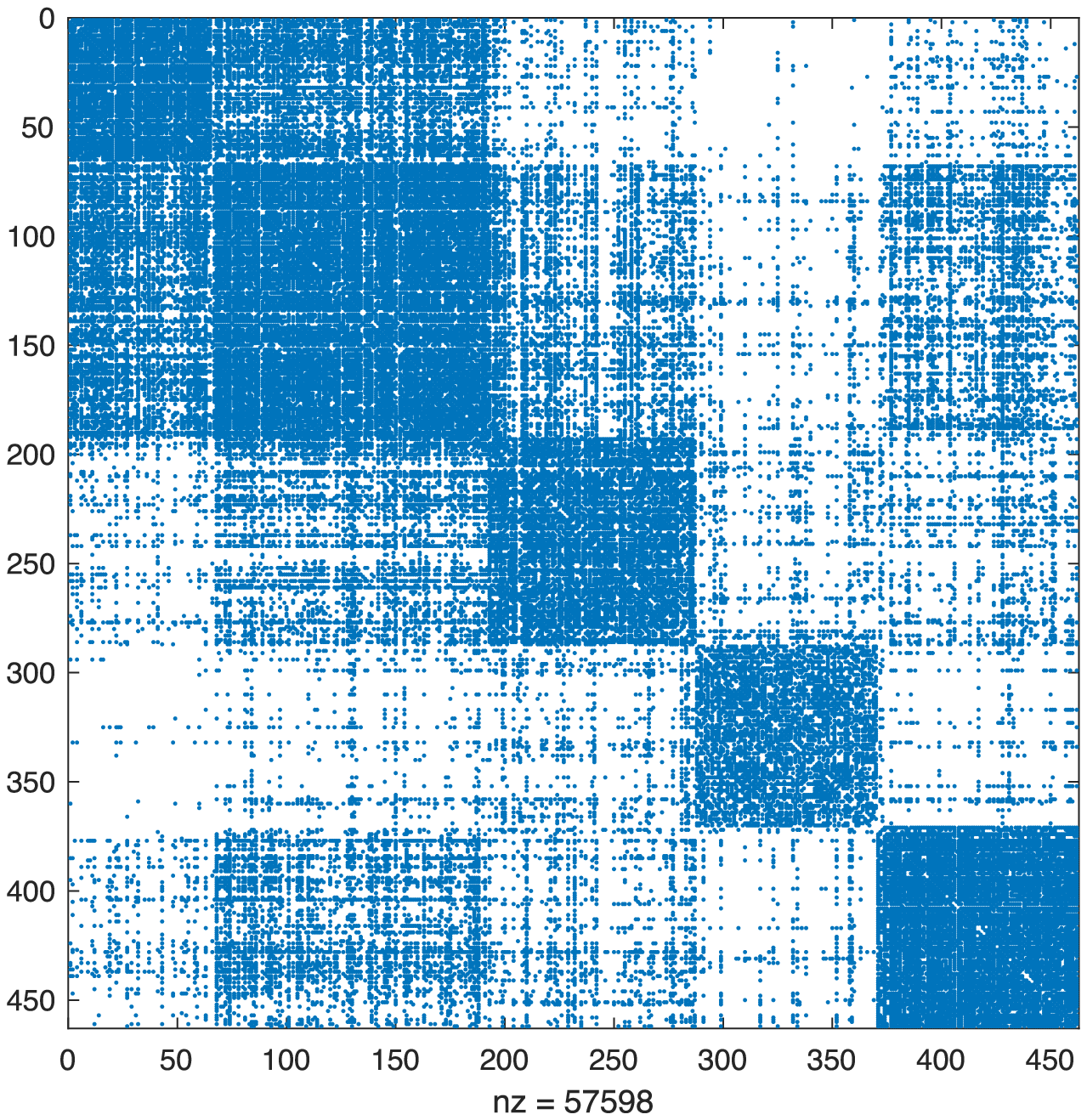} \hspace{10mm} 
\includegraphics[height=7cm]{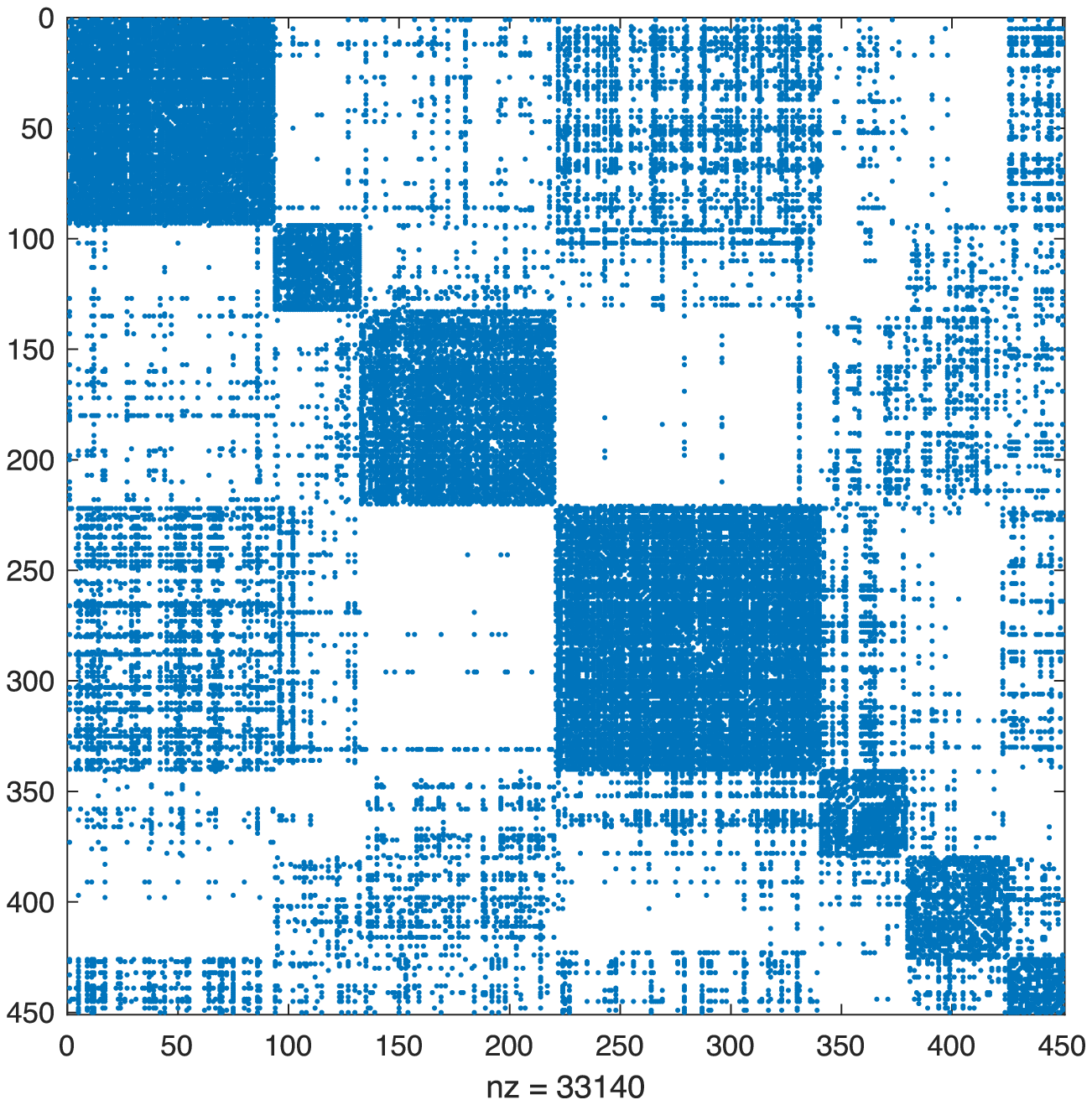} \]  
  \caption{The adjacency matrices of the butterfly similarity network with 57598 nonzero entries and 5 clusters (left) and the brain network with 33140 nonzero entries and 7 clusters (right) after clustering}
   \label{mn:fig_Adjacency matrices}
   \end{center}
\end{figure}

In this section, we describe application of the two-step clustering procedure of Section~\ref{sec:clustering}  to two real life networks, a butterfly similarity network and a human brain network. 

We consider  the butterfly similarity network extracted from the Leeds Butterfly  dataset 
\cite{wang2018network}, which contains fine-grained  images of 832 butterfly species 
that belong to 10 different classes, with each class containing between 55 and 100 images. 
In this network, the nodes represent butterfly species and edges represent visual similarities 
(ranging from 0 to 1) between them, evaluated on the basis of butterfly images.
We extract  the five largest classes and draw an edge between two nodes if  the visual 
similarity between them is greater than zero, obtaining  a simple graph  with  462 nodes and 28799 edges.
We carry out clustering of the nodes, employing the two-step clustering procedure, 
first finding $L=4$ meta-communities by Algorithm~\ref{mn:SSC:alg},  and then using Algorithm~\ref{mn:SC_k_median:alg} to find communities within meta-communities. 
We conclude that the first meta-community has two communities,
while the other three meta-communities have one community each. 
We also  applied Algorithms~\ref{mn:SSC:alg}~and~\ref{mn:SC_k_median:alg} 
separately  for detection of five communities.
Here,  Algorithms~\ref{mn:SSC:alg}~and~\ref{mn:SC_k_median:alg} correspond, respectively, 
to the PABM and the DCBM settings with $K=5$.  
Subsequently, we compare  the clustering assignments 
with the true class specifications of the species.
Algorithms~\ref{mn:SSC:alg}~and~\ref{mn:SC_k_median:alg}  lead to 74\% and 77\% accuracy, respectively,
while the  two-step clustering procedure provides better 84\% accuracy, thus,  
justifying the application of the NBM. The better results are due to the higher flexibility of the NBM.

 \ignore{
Our results show that the two-step clustering procedure provides a better accuracy than Algorithm~\ref{mn:SSC:alg} and Algorithm~\ref{mn:SC_k_median:alg}. 
Figure \ref{mn:fig_Adjacency matrices} (left) shows the adjacency matrix of the graph (after clustering) using the two-step clustering procedure which confirms that the first two communities form a meta-community.
Here we see that the NBM allowing some of the butterflies in one class to be ''more similar'' in some aspects to species of another specific class than the other, meaning those classes can form a meta-community, thus, justifying the application of the NBM.
}

The second example deals with analysis of a human brain functional network, 
based on the brain connectivity dataset, derived from the resting-state functional 
MRI (rsfMRI) \cite{crossley2013cognitive}. 
%
In this dataset, the brain is partitioned into 638 distinct regions and a weighted graph is used to characterize the network topology. 
For a comparison, we use the Asymptotical Surprise method \cite{nicolini2017community} 
which is applied for clustering the GroupAverage rsfMRI matrix in \cite{crossley2013cognitive}. 
Asymptotical Surprise detects 47 communities with sizes ranging from 1 to 133. 
Since the true clustering as well as the true number of clusters are unknown
for this dataset, we treat the results of the Asymptotical Surprise as the ground truth.
In order to generate a binary network, we set all nonzero weights to one in the GroupAverage rsfMRI matrix, 
obtaining a network with 18625 undirected edges.  For our study, we extract 7 largest 
communities derived by the Asymptotical Surprise, obtaining  a   network with 450 nodes and 16570 edges. 
Similarly to the previous example, we apply Algorithms~\ref{mn:SSC:alg}~and~\ref{mn:SC_k_median:alg} 
separately to detect seven communities, obtaining,  respectively, 88\% and 73\% accuracy.
We also use  the the two-step clustering procedure above, detecting six meta-communities and
seven communities, attaining  92\% accuracy.

Figure \ref{mn:fig_Adjacency matrices}  shows the adjacency matrices
of the butterfly similarity network (left) and the human brain network (right)
 after clustering.


\section {Discussion}
\label{sec:discussion}

The present paper examines the hierarchy of block models with the purpose 
of treating all existing singular-membership block models as a part of one formulation,
which is free from arbitrary identifiability conditions. 
 The blocks differ by the average probability of connections and can be combined into
  meta-blocks that have common heterogeneity patterns in the connection probabilities. 
 
The hierarchical formulation proposed above  
(see Figure~\ref{mn:fig01}) can be utilized for a variety of purposes.
Since the NBM treats all other block models as its particular cases,
one can carry out estimation and clustering without assuming that a specific block model holds, by employing the NBM with $K$
communities and $L$ meta-communities, where both 
$K$ and $L$ are unknown. 
The values of $K$ and $L$ can later be derived on the basis of penalties.
Furthermore, in the framework above, one can easily test one block model versus another. 
For instance, $L=K$ suggests the PABM while $L=1$ implies the DCBM. If,   additionally,
$H = 1_n$, then DCBM reduces to SBM. Finally, one can see from Figure~\ref{mn:fig01} that
the absence of distinct communities ($K=1$) always leads to DCBM, which reduces to 
Erd\H{o}s-R\'{e}nyi model if  $H = 1_n$. 









\section {Proofs}
\label{sec:proofs}



\subsection{Proof  of Theorem~\ref{th:oracle_Klopp}. }
Let $\Xi = A- P_*$. We let $\mathscr{P}_{Z, C, K, L}$ denote the permutation matrix that
arranges meta-blocks consecutively and also blocks  all meta-blocks consecutively. 
For simplicity, let
$$\mathscr{P} \equiv \mathscr{P}_{Z, C, K, L}, \hspace{1mm} \mathscr{P}_* \equiv \mathscr{P}_{Z_*, C_*, K_*, L_*}, 
\hspace{1mm} \hat{\mathscr{P}} \equiv \mathscr{P}_{\hat{Z}, \hat{C}, \hat{K}, \hat{L}}$$
For any matrix $S$, denote
\be \label{HBM:eq:permute}
S(Z,C,K,L)=\mathscr{P}_{Z, C, K, L}^TS\mathscr{P}_{Z, C, K, L}
\ee
Then, for any $Z,C,K,$ and  $L$:
\bes
\begin{aligned}
 \left\| \hat{\mathscr{P}}^T A\hat{\mathscr{P}}-\hat\Theta(\hat{Z},\hat{C},\hat{K},\hat{L}) \right\|_F^2 + \Pen(n,\hat{K},\hat{L}) 
\leq  \left\| \mathscr{P}^T A \mathscr{P}  - \mathscr{P}^T P  \mathscr{P} \right\|_F^2 + 
 \Pen(n,K,L) 
\end{aligned}
\ees
Therefore, 
\bes 
\begin{aligned}
\left\| A - \hat{\mathscr{P}}\hat\Theta(\hat{Z}, \hat{C},\hat{K}, \hat{L}) \hat{\mathscr{P}}^T \right\| _F^2 
 + \Pen(n,\hat{K},\hat{L})\leq  \left\| A - P \right \| _F^2  + \Pen(n,K,L)
\end{aligned}
\ees 
or
\be \label{eq:main_ineq0_klopp}
\begin{aligned}
\left\| A - \hat{P} \right\| _F^2 
 + \Pen(n,\hat{K},\hat{L})\leq  \left\| A - P \right\| _F^2  + \Pen(n,K,L).
\end{aligned}
\ee 
Subtracting and adding $P_*$ in the norms in both sides of  \eqref{eq:main_ineq0_klopp},
we rewrite it as 
\be \label{eq:tot_err_klopp}
\begin{aligned}
  \left\|\hat{P}  -P_{*}\right\|_F^2 \leq  \left\|P  -P_{*}\right\|_F^2  + 2\langle\Xi, \hat{P} - P \rangle +   \Pen(n,K,L) - \Pen(n,\hat{K},\hat{L}).
  \end{aligned}
\ee
Denote
$$P_0(K,L)=\underset{P \in \Im(n,K,L)}{\text{inf}} \left\| P  -P_{*}\right\| _F^2,$$
$$(K_0,L_0)=\underset{K,L}{\text{inf}} \left \{ \left\| P_0(K,L)  -P_{*}\right\| _F^2 + \Pen(n,K,L) \right \}.$$
Then, for $\hat{P} \equiv \hat{P}(\hat{K},\hat{L})$ and $P_0 \equiv P_0(K_0,L_0)$, one has
\be 
\begin{aligned}  \label{eq:main_err_klopp}
&  \left\| \hat{P}  -P_{*}\right\| _F^2 \leq  \left\| P_0  -P_{*}\right\| _F^2  +  2\langle\Xi, P_* - P_0 \rangle \\  & 2\langle\Xi, \hat{P} - P_* \rangle + \Pen(n,K_0,L_0) - \Pen(n,\hat{K},\hat{L}).
  \end{aligned}
\ee
Denote
\be \label{eq:taunkl}
\begin{aligned} 
& \tau(n,K,L) = 
& n \ln K + K \ln L + (K^2 + 2 n L ) \ln \left( 9 n L \right)
\end{aligned}
\ee 
and consider two sets $\Omega$ and $\Omega^c$ 
\be
\begin{aligned} \label{eq:omega_sets}
\Omega = \left \{ \omega :  \left\| \hat{P}  -P_{*}\right\| _F \geq C_0 2^{s_0} \sqrt{\tau(n,K_0,L_0) } \right \}, \\
\Omega^c = \left \{ \omega :  \left\| \hat{P}  -P_{*}\right\| _F \leq C_0 2^{s_0} \sqrt{\tau(n,K_0,L_0) } \right \}
\end{aligned}
\ee
where $s_0$ is a constant. If $\omega \in \Omega^c$, then
\be
\begin{aligned} \label{eq:omega_c}
 \left\| \hat{P}  -P_{*}\right\| _F^2 \leq C_0^2 2^{2s_0} \tau(n,K_0,L_0) 
\end{aligned}
\ee
Consider the case when $\omega \in \Omega$. It follows from Hoeffding inequality that,  for any
fixed matrix $G$, any $\alpha>0$ and any $t>0$ one has  
\be \label{eq:aux_lemma}
\PP \lfi 2\langle\Xi, G \rangle \geq \alpha \left\|G\right\|_F^2 +  2t/\alpha \rfi \leq e^{-t}.
\ee
Then, there exists a set $\tilde\Omega_t$ such that $P(\tilde{\Omega}_t) \geq 1 - e^{-t}$ 
and for $\omega \in \tilde\Omega_t$
\be
\begin{aligned} \label{eq:up_bound_1}
 2\langle\Xi, P_* - P_0 \rangle \leq \alpha \left\| P_*  -P_0 \right\| _F^2 +  2t/\alpha 
\end{aligned}
\ee
Note that the set $\Omega$ can be partitioned as 
$\Omega = \underset{K,L} {\bigcup} \Omega_{K,L}$, where
\be \label{eq:Omkl}
\begin{aligned} 
 \Omega_{K,L} = \bigg \{ \omega : & \Big( \left\| \hat{P}  -P_{*} \right\| _F \geq C_0 2^{s_0} \sqrt{\tau(n,K_0,L_0) } \Big)   \hspace{1mm}
 \cap \hspace{1mm} (\hat{K}=K, \hspace{1mm} \hat{L}=L)  \bigg \}
\end{aligned}
\ee 
 with $\Omega_{K_1,L_1} \cap  \Omega_{K_2,L_2} = \emptyset$ 
 unless $K_1 = K_2$ and $L_1 = L_2$.
Denote
\be \label{eq:DelnKL}
\Delta(n,K,L) = C_0^2 C_2 \tau (n, K, L) +n,
\ee
where $\tau (n, K, L)$  is defined in \eqref{eq:taunkl}.
Then,  
\bes
\begin{aligned} 
&\PP \bigg \{ \Big [ 2\langle\Xi, \hat{P}(n,\hat{K}, \hat{L}) - P_* \rangle -  
 \frac{1}{2} \left\| \hat{P}(n,\hat{K}, \hat{L}) - P_{*}\right\| _F^2 -
2 \Delta(n,\hat{K}, \hat{L}) \Big ] \geq 0  \bigg \}\\
& \leq \sum_{K=1}^{n} \sum_{L=1}^{K}  \PP \bigg \{ \underset{\hat{P} \in \Omega_{K,L}}{\text{sup}} \Big [ 2\langle\Xi, \hat{P} - P_* \rangle -  
 \frac{1}{2} \left\| \hat{P} - P_{*} \right\| _F^2 - 
 2 \Delta(n, K, L) \Big ] \geq 0  \bigg \}
\end{aligned}
\ees
By Lemma~\ref{lem:upper_bound_klopp} in Section
\ref{sec:Suppl_statements}, there exist sets 
$\tilde{\Omega}_{K,L} \subseteq \Omega_{K,L} \subset \Omega$ such that
$\PP(\tilde{\Omega}_{K,L}^c) \leq \log_2 n \cdot \exp{(-n \cdot 2^{2s_0 - 7})}$ 
and, for $\omega \in \tilde{\Omega}_{K,L}$, one has
\bes
\begin{aligned} 
\Big \{ 2\langle\Xi, \hat{P} - P_* \rangle \leq 
\frac{1}{2} \left\| \hat{P} - P_{*}\right\| _F^2 + 2 \Delta(n, K, L) \Big \} \cap  \Big \{ \hat{K}=K, \hspace{1mm} \hat{L}=L \Big \} 
\end{aligned}
\ees
Denote 
\be \label{eq:tol_Omega}
\tilde{\Omega} = \left ( \underset{K,L} {\cap} \tilde{\Omega}_{K,L} \right) 
\cap \tilde{\Omega}_t 
\ee
and observe that 
\bes
 \PP  \lkr  \tilde{\Omega}  \rkr \geq  1- 
  n^2 \log_2 n \cdot \exp{(-n \cdot 2^{2s_0 - 7})} - e^{-t}.
\ees
Then, for $\omega \in \tilde{\Omega}$, one has
\be \label{eq:up_bound_2}
\begin{aligned} 
 2\langle\Xi, \hat{P} - P_* \rangle \leq 
\frac{1}{2} \left\| \hat{P} - P_{*} \right\| _F^2 + 2 \Delta(n, \hat{K}, \hat{L}) 
\end{aligned}
\ee
and it follows from (\ref{eq:up_bound_1}) with $\alpha = 1/2$ that
\be \label{eq:up_bound_3}
\begin{aligned} 
 2\langle\Xi, P_* - P_0 \rangle \leq 
\frac{1}{2} \left\| P_* - P_0 \right\| _F^2 + 4 t 
\end{aligned}
\ee
Plugging (\ref{eq:up_bound_2}) and (\ref{eq:up_bound_3}) into (\ref{eq:main_err_klopp}),
obtain that  for $\om \in \tilde{\Omega}$ one has
\bes 
\begin{aligned}  
  \left\| \hat{P}  -P_{*}\right\| _F^2 \leq  \left\| P_0  -P_{*}\right\| _F^2  +  \Pen(n,K_0,L_0) + 
  \frac{1}{2} \left\| \hat{P}  - P_{*}\right\| _F^2  + \\   2 \Delta(n,\hat{K}, \hat{L}) + \frac{1}{2} \left\| P_* - P_0\right\| _F^2  + 
 4 t - \Pen(n,\hat{K},\hat{L})
\end{aligned}
\ees
Finally, setting
\bes 
\begin{aligned}  
  \Pen(n,K,L) = 2 \Delta (n,K,L)  = 2 \lkv  C_0^2 \tau (n,K,L) + n \rkv,
\end{aligned}
\ees
obtain that for any $t > 0$, for  $\om \in \tilde{\Omega}$,
one has
\bes 
\begin{aligned}  
  \left\| \hat{P}  -P_{*}\right\| _F^2 \leq  3 \left\| P_0  -P_{*}\right\| _F^2  +  2 \Pen(n,K_0,L_0) + 8 t,
\end{aligned}
\ees
for any $\omega \in \Omega$.
Now, for $\omega \in \Omega^c$, it follows  from (\ref{eq:omega_c}) that 
\bes 
\begin{aligned}  
  \left\| \hat{P}  -P_{*}\right\| _F^2 \leq C_0^2 2^{2s_0} \tau(n, K_0, L_0) \leq 
  2^{2 s_0 -1} \Pen(n,K_0,L_0)
\end{aligned}
\ees
Setting $s_0=1$ and $t = n/32$, obtain
\bes 
\begin{aligned}  
 \PP \bigg \{  \left\| \hat{P}  -P_{*}\right\| _F^2 \leq \Big [ 3 \left\| P_0 - P_* \right\| _F^2 + 2 \Pen(n,K_0,L_0) \Big ] + 
 \frac{n}{4} \bigg \} \geq 
 1 - (n^2 \log_2 n +1) e^{- \frac{n}{32}},
\end{aligned}
\ees
so that 
\bes 
\begin{aligned}  
 \PP \bigg \{  \left\| \hat{P}  -P_{*} \right\| _F^2 \leq 3 \underset{P \in  \Im(n,K,L)}{\text{inf}} \Big[  \left\| P - P_* \right\| _F^2 + 
 \Pen(n,K,L) \Big ] \bigg \}  \geq 
 1 - (n^2 \log_2 n +1) e^{- \frac{n}{32}}
\end{aligned}
\ees
Since $\left\| \hat{P}  -P_{*}\right\| _F^2 \leq n^2$, obtain
\bes 
\begin{aligned}  
 \EE  \left\| \hat{P}  -P_{*} \right\| _F^2 \leq 3 \underset{P \in  \mathscr{M}(n,K,L)}{\text{min}} \Big[  \left\| P - P_* \right\| _F^2 + 
 \Pen(n,K,L) \Big ] + n^5 e^{- n/32}
\end{aligned}
\ees

\subsection{Proof  of Theorem~\ref{th:oracle}. }
Let
\bes 
\begin{aligned}
& F_1(n,K,L)= 
 C_1n K + C_2 K^2 \ln (ne) +C_3( \ln n + (n+1) \ln K + K \ln L)\\
& F_2(n,K,L)=2 \ln n + 2(n+1) \ln K + 2K \ln L,
\end{aligned}
\ees
where $C_1,$ $C_2,$ and $C_3$  are  absolute constants. 
Denote $\Xi = A- P_*$ and recall that,  given matrix $P_*$, entries  
$\Xi_{i,j} = A_{i,j}-(P_{*})_{ij}$ of $\Xi$  are the independent Bernoulli errors for $1 \leq i \leq j \leq n$ and
$ A_{i,j}= A_{j,i}$. Then, following notation \eqref{HBM:eq:permute}, for any $Z$, $C$, $K$, and $L$
\bes 
\begin{aligned}
\Xi(Z,C,K,L) = \mathscr{P}^T\Xi \mathscr{P}\\ 
 P_{*} (Z,C,K,L) = \mathscr{P}^T P_{*} \mathscr{P},
\end{aligned}
\ees
where $\mathscr{P} \equiv \mathscr{P}_{Z, C, K, L}$. Then it follows from \eqref{eq:opt_main} that 
\bes
\begin{aligned}
\left\|\hat{\mathscr{P}}^T A\hat{\mathscr{P}}-\hat\Theta(\hat{Z},\hat{C},\hat{K},\hat{L}) \right\|_F^2 + \Pen(n,\hat{K},\hat{L}) 
\leq  \left\|\mathscr{P}_{*}^T A \mathscr{P}_{*}  - \mathscr{P}_{*}^T P_{*}  \mathscr{P}_{*}\right\|_F^2 +  \Pen(n,K_{*},L_{*}) 
\end{aligned}
\ees
where $\mathscr{P}_{*} \equiv \mathscr{P}_{Z_*, C_*, K_*, L_*}$. Using the fact that permutation matrices are orthogonal, we can rewrite the previous inequality as 
\be \label{eq:main_ineq0}
\begin{aligned}
\left\| A - \hat{\mathscr{P}}\hat\Theta(\hat{Z}, \hat{C},\hat{K}, \hat{L}) \hat{\mathscr{P}}^T \right\|_F^2 
 + \Pen(n,\hat{K},\hat{L})\leq  \left\|A - P_{*}\right\|_F^2  + \Pen(n,K_{*},L_{*}).
\end{aligned}
\ee 
Hence,  \eqref{eq:main_ineq0} and  \eqref{eq:P_total_est}  yield 
\be  \label{eq:main_ineq1}
\norm{ A - \hat{P}}_F^2 \leq \norm{A - P_{*}}_F^2  + \Pen(n,K_{*},L_{*}) - \Pen(n,\hat{K},\hat{L}) 
\ee
Subtracting and adding $P_*$ in the norm of the left-hand side of  \eqref{eq:main_ineq1},
we rewrite \eqref{eq:main_ineq1} as 
\be \label{eq:tot_err}
\begin{aligned}
  \left\|\hat{P}  -P_{*}\right\|_F^2 \leq  \Delta(\hat{Z}, \hat{C},\hat{K}, \hat{L}) +  \Pen(n,K_{*},L_{*}) - \Pen(n,\hat{K},\hat{L}), \end{aligned}
\ee
where 
\be \label{eq:DelZK}
  \Delta \equiv \Delta (\hat{Z}, \hat{C},\hat{K}, \hat{L}) = 2 \Tr\left[\Xi^T (\hat{P}  -P_{*})\right].
\ee

Again, using orthogonality of the permutation matrices, we can rewrite 
\bes
\begin{aligned}
\Delta = 2 \langle\Xi(\hat{Z},\hat{C},\hat{K},\hat{L}),(\hat\Theta(\hat{Z},\hat{C},\hat{K},\hat{L}) - P_{*} (\hat{Z},\hat{C},\hat{K},\hat{L}))\rangle,
\end{aligned}
\ees
where $\langle A,B\rangle = \Tr(A^TB)$.
Then, in the block form, $\Delta$ appears as 
\be \label{eq:Del_blockform}
\Delta =  \displaystyle \sum_{l=1}^{\hat{L}} \sum_{k=1}^{\hat{K}} \Delta^{(l,k)}
\ee
where 
\bes 
\begin{aligned}
\Delta^{(l,k)} =  2 \langle\Xi^{(l,k)}(\hat{Z},\hat{C},\hat{K},\hat{L}), \Pi_{\hat{u}, \hat{v}}( A^{(l,k)}(\hat{Z},\hat{C},\hat{K},\hat{L})) - 
P_{*}^{(l,k)} (\hat{Z},\hat{C},\hat{K},\hat{L})
\rangle 
\end{aligned}
\ees 
and $\Pi_{\hat{u}, \hat{v}}$ is defined in   \eqref{eq:Pi_uv} of Lemma~\ref{lem:lowrank_approx}.

Let $\tilde {u}=\tilde {u}^{(l,k)}(\hat{Z},\hat{C},\hat{K},\hat{L})$ and $\tilde {v}={\tilde{v}^{(l,k)}}(\hat{Z},\hat{C},\hat{K},\hat{L})$ be the singular vectors 
of $P_{*}^{(l,k)}(\hat{Z},\hat{C},\hat{K},\hat{L})$ corresponding to the largest singular value  of 
$P_{*}^{(l,k)}(\hat{Z},\hat{C},\hat{K},\hat{L})$. 
Then, according to Lemma~\ref{lem:lowrank_approx}
\be \label{not_r1approx_kl_block of P}
\begin{aligned}
 &\Pi_{\tilde{u}, \tilde{v}} \left(P_{*}^{(l,k)}(\hat{Z},\hat{C},\hat{K},\hat{L})\right) = 
&\tilde {u}^{(l,k)}(\tilde{u}^{(l,k)})^T P_{*}^{(l,k)} \tilde{v}^{(l,k)} (\tilde{v}^{(l,k)} )^T
\end{aligned}
\ee
Recall that 
\bes
\begin{aligned}
 &\Pi_{\hat{u}, \hat{v}} (A^{(l,k)}(\hat{Z},\hat{C},\hat{K},\hat{L})) = &\Pi_{\hat{u},\hat{v}}  
\left[P_{*}^{(l,k)} (\hat{Z},\hat{C},\hat{K},\hat{L}) + \Xi^{(l,k)}(\hat{Z},\hat{C},\hat{K},\hat{L})\right], 
\end{aligned}
\ees
Then, $\Delta^{(l,k)}$ can be partitioned into the sums of three components
\be \label{eq:Delkl_sum}
\begin{aligned}
&\Delta^{(l,k)} = \Delta_1^{(l,k)} + \Delta_2^{(l,k)} + 
\Delta_3^{(l,k)}, \quad 
&l = 1,2,\cdots, \hat{L}, \hspace{2 mm} k = 1,2,\cdots, \hat{K}
\end{aligned}
\ee
where 

\be \label{kl_blockdelta1_main_est_error}
\begin{aligned}
\Delta_1^{(l,k)}  =  2 \langle \Xi^{(l,k)}(\hat{Z},\hat{C},\hat{K},\hat{L}),\Pi_{\hat{u}, \hat{v}} 
( \Xi^{(l,k)}(\hat{Z},\hat{C},\hat{K},\hat{L})) \rangle
\end{aligned}
\ee

\be \label{kl_blockdelta2_main_est_error}
\begin{aligned}
 \Delta_2^{(l,k)}  =   2 \langle \Xi^{(l,k)}(\hat{Z},\hat{C},\hat{K},\hat{L}), \Pi_{\tilde{u}, \tilde{v}} 
(P_{*}^{(l,k)}(\hat{Z},\hat{C},\hat{K},\hat{L})) 
 - P_{*} ^{(l,k)}(\hat{Z},\hat{C},\hat{K},\hat{L})\rangle
\end{aligned}
\ee

\be \label{kl_blockdelta3_main_est_error}
\begin{aligned}
 \Delta_3^{(l,k)}  =  2\langle \Xi^{(l,k)}(\hat{Z},\hat{C},\hat{K},\hat{L}),\Pi_{\hat{u}, \hat{v}} 
(P_{*}^{(l,k)}(\hat{Z},\hat{C},\hat{K},\hat{L}) )  
 - 
 \Pi_{\tilde{u}, \tilde{v}} (P_{*}^{(l,k)}(\hat{Z},\hat{C},\hat{K},\hat{L})) \rangle 
\end{aligned}
\ee

\ignore{
\beqn 
\label{kl_blockdelta1_main_est_error}
\begin{small}
\Delta_1^{(l,k)} & = &  2 \langle \Xi^{(l,k)}(\hat{Z},\hat{C},\hat{K},\hat{L}),\Pi_{\hat{u}, \hat{v}} 
( \Xi^{(l,k)}(\hat{Z},\hat{C},\hat{K},\hat{L})) \rangle
\\
\label{kl_blockdelta2_main_est_error}
 \Delta_2^{(l,k)} & = &  2 \langle \Xi^{(l,k)}(\hat{Z},\hat{C},\hat{K},\hat{L}), \Pi_{\tilde{u}, \tilde{v}} 
\left(P_{*}^{(l,k)}(\hat{Z},\hat{C},\hat{K},\hat{L})\right)- P_{*} ^{(l,k)}(\hat{Z},\hat{C},\hat{K},\hat{L})\rangle\\
\label{kl_blockdelta3_main_est_error}
 \Delta_3^{(l,k)} & = & 2\langle \Xi^{(l,k)}(\hat{Z},\hat{C},\hat{K},\hat{L}),\Pi_{\hat{u}, \hat{v}} 
(P_{*}^{(l,k)}(\hat{Z},\hat{C},\hat{K},\hat{L}) ) -
\Pi_{\tilde{u}, \tilde{v}} \left(P_{*}^{(l,k)}(\hat{Z},\hat{C},\hat{K},\hat{L})\right) \rangle 
\end{small}
\eeqn
}

With some abuse of notations, for any matrix $B$, let $\Pi_{\tilde{u}, \tilde{v}} \left(B(\hat{Z},\hat{C},\hat{K},\hat{L})\right)$ 
be  the matrix with blocks $\Pi_{\tilde{u}, \tilde{v}} \left(B^{(l,k)}(\hat{Z},\hat{C},\hat{K},\hat{L})\right)$ and 
$\Pi_{\hat{u}, \hat{v}}  \left(B(\hat{Z},\hat{C},\hat{K},\hat{L})\right)$ be  the matrix with blocks 
$$\Pi_{\hat{u}, \hat{v}} \left(B^{(l,k)}(\hat{Z},\hat{C},\hat{K},\hat{L})\right),
\hspace{1mm} l = 1,2,\cdots, \hat{L}, \hspace{1mm} k = 1,2,\cdots, \hat{K}. $$ 
Then, it follows from \eqref{eq:Delkl_sum}--\eqref{kl_blockdelta3_main_est_error} that 
\be \label{eq:Del_sum}
\Delta = \Delta_1 + \Delta_2 + \Delta_3  
\ee
where

\be \label{delta1_main_est_error}
\begin{aligned}
 \Delta_1  =  2 \langle (\Xi(\hat{Z},\hat{C},\hat{K},\hat{L}),\Pi_{\hat{u}, \hat{v}} ( \Xi(\hat{Z},\hat{C},\hat{K},\hat{L})) \rangle
\end{aligned}
\ee

\be  \label{delta2_main_est_error}
\begin{aligned}
 \Delta_2  =   2 \langle \Xi(\hat{Z},\hat{K}), \Pi_{\tilde{u}, \tilde{v}}( P_{*}(\hat{Z},\hat{C},\hat{K},\hat{L}))-  
 P_{*} (\hat{Z},\hat{C},\hat{K},\hat{L})\rangle
\end{aligned}
\ee

\be \label{delta3_main_est_error}
\begin{aligned}
 \Delta_3  =   2\langle \Xi(\hat{Z},\hat{C},\hat{K},\hat{L}),\Pi_{\hat{u}, \hat{v}} (P_{*}(\hat{Z},\hat{C},\hat{K},\hat{L}) ) - 
\Pi_{\tilde{u}, \tilde{v}}( P_{*}(\hat{Z},\hat{C},\hat{K},\hat{L})) \rangle 
\end{aligned}
\ee

\ignore{

\beqn 
\label{delta1_main_est_error}
\Delta_1 & = & 2 \langle (\Xi(\hat{Z},\hat{C},\hat{K},\hat{L}),\Pi_{\hat{u}, \hat{v}} ( \Xi(\hat{Z},\hat{C},\hat{K},\hat{L})) \rangle \\
\label{delta2_main_est_error}
 \Delta_2 & = &  2 \langle \Xi(\hat{Z},\hat{K}), \Pi_{\tilde{u}, \tilde{v}}\left( P_{*}(\hat{Z},\hat{C},\hat{K},\hat{L})\right)- 
P_{*} (\hat{Z},\hat{C},\hat{K},\hat{L})\rangle\\
\label{delta3_main_est_error}
 \Delta_3 & = &  2\langle \Xi(\hat{Z},\hat{C},\hat{K},\hat{L}),\Pi_{\hat{u}, \hat{v}} (P_{*}(\hat{Z},\hat{C},\hat{K},\hat{L}) ) -
\Pi_{\tilde{u}, \tilde{v}}\left( P_{*}(\hat{Z},\hat{C},\hat{K},\hat{L})\right) \rangle  
\eeqn
Now, we need to derive an upper bound for each component in \eqref{eq:Delkl_sum} and \eqref{eq:Del_sum}.
\ignore{
Let 
\beqn \label{eq:F1}
F_1(n,K) & = &  72 nK +  2K^2 \tilde{c} \left[(\tau+ 2)\ln n + (n+1) \ln K \right],\\
\label{eq:F2}
F_2(\tau,n,K) & = &  2((\tau +1)\ln n + n \ln K).
\eeqn
}
}


Observe that 
\begin{align*}
\Delta_1^{(l,k)}&   
= 2 \langle \Xi^{(l,k)}(\hat{Z},\hat{C},\hat{K},\hat{L}),\Pi_{\hat{u}, \hat{v}} ( \Xi^{(l,k)}(\hat{Z},\hat{C},\hat{K},\hat{L})) \rangle  \\
&= 2 \left\|\Pi_{\hat{u}, \hat{v}} ( \Xi^{(l,k)}(\hat{Z},\hat{C},\hat{K},\hat{L}))\right\|_F^2\\
& \leq 2 \left\| \Xi^{(l,k)}(\hat{Z},\hat{C},\hat{K},\hat{L}) \right\|_{op}^2.  
\end{align*}  
Now, fix $t$ and let $\Omega_{1}$ be the set where $$\di\sum_{l=1}^{\hat{L}} \sum_{k=1}^{\hat{K}} \left\|\Xi^{(l,k)} (\hat{Z},\hat{C},\hat{K},\hat{L}) \right\|_{op}^2 \leq  F_1(n,\hat{K},\hat{L}) + C_3 t$$.
According to Lemma~\ref{lem:prob_error_bound_main}, 
\be \label{eq:P_Omega1}
\PP(\Omega_{1}) \geq  1-   \exp(-t),
\ee
and,  for $\omega \in \Omega_{1}$, one has 
\be \label{eq:Del1_bound}
\begin{aligned}
& \vert \Delta_1\vert \leq  
  2 \di\sum_{l=1}^{\hat{L}} \sum_{k=1}^{\hat{K}} \left\|\Xi^{(l,k)}(\hat{Z},\hat{C},\hat{K},\hat{L})\right\|_{op}^2 
  & \leq 2 F_1(n,\hat{K},\hat{L}) + 2 C_3 t
\end{aligned}
\ee


Now, consider  $\Delta_2$ given by \eqref{delta2_main_est_error}.
Note that 
\be \label{eqn_useful}
\begin{aligned}
| \Delta_2 | =
2 \left\|\Pi_{\tilde{u}, \tilde{v}}\left( P_{*}(\hat{Z},\hat{C},\hat{K},\hat{L})\right)- 
P_{*} (\hat{Z},\hat{C},\hat{K},\hat{L}) \right\|_F 
| \langle \Xi(\hat{Z},\hat{C},\hat{K},\hat{L}),H_{\tilde{u}, \tilde{v}}(\hat{Z},\hat{C},\hat{K},\hat{L}) \rangle|
\end{aligned}
\ee  
where 
\bes
\begin{aligned}
& H_{\tilde{u},\tilde{v}}(\hat{Z},\hat{C},\hat{K},\hat{L})= 
& \frac{\Pi_{\tilde{u}, \tilde{v}}\left( P_{*}(\hat{Z},\hat{C},\hat{K},\hat{L})\right)- 
P_{*} (\hat{Z},\hat{C},\hat{K},\hat{L})}{\|\Pi_{\tilde{u},\tilde{v}}\left( P_{*}(\hat{Z},\hat{C},\hat{K},\hat{L})\right)- P_{*} (\hat{Z},\hat{C},\hat{K},\hat{L})\|_F }
\end{aligned}
\ees
Since  for any $a$, $b$, and $\alpha_1 > 0$, one has $2ab \leq \alpha_1 a^2 +  b^2/\alpha_1$, obtain
\be \label{eq:Del2_sum} 
\begin{aligned}
& |\Delta_2| \leq  \alpha_1 \left\|\Pi_{\tilde{u}, \tilde{v}}\left( P_{*}(\hat{Z},\hat{C},\hat{K},\hat{L})\right)- 
P_{*} (\hat{Z},\hat{C},\hat{K},\hat{L}) \right\|_F^2 + \\
& 1/\alpha_1\, |  \langle \, \Xi(\hat{Z},\hat{C},\hat{K},\hat{L}),H_{\tilde{u},\tilde{v}}(\hat{Z},\hat{C},\hat{K},\hat{L})\, \rangle|^2
\end{aligned}
\ee 
Observe that if $K$, $L$, $Z \in \calM_{n,K}$, and $C \in \calM_{K,L}$ are fixed, then $H_{\tilde{u},\tilde{v}}(Z,C,K,L)$ 
is  fixed and, for any $K$, $L$, $Z$, and $C$, one has  $\| H_{\tilde{u},\tilde{v}}(Z,C,K,L)\|_F =1$.
Note also that, for fixed $K$, $L$, $Z$, and $C$, permuted matrix  $\Xi(Z,C,K,L) \in [0,1]^{n\times n}$ 
contains independent Bernoulli errors. It is well known that if $\xi$ is a vector of independent Bernoulli errors
and $h$ is a unit vector, then, for any $x>0$,   Hoeffding's inequality yields
$$ 
\PP(|\xi^T h|^2> x) \leq 2 \exp(- x/2) 
$$
Since 
\bes
\begin{aligned}
\langle \Xi(Z,C,K,L),H_{\tilde{u},\tilde{v}}(Z,C,K,L)\rangle = 
[\vect(\Xi(Z,C,K,L))]^T  \vect(H_{\tilde{u},\tilde{v}}(Z,C,K,L)),
\end{aligned}
\ees
obtain  for any fixed $K$, $L$, $Z$, and $C$:
\bes 
\begin{aligned} 
\PP  \left( | \langle \Xi(Z,C,K,L),H_{\tilde{u},\tilde{v}}(Z,C,K,L)\rangle|^2 - x > 0 \right) \leq  2 \exp(-x/2) 
\end{aligned}
\ees
Now, applying the union bound, derive
\begin{align} 
& \PP  \left( | \langle \Xi(\hat{Z},\hat{C},\hat{K},\hat{L}),H_{\tilde{u},\tilde{v}}(\hat{Z},\hat{C},\hat{K},\hat{L}) \rangle|^2 - F_2(n,\hat{K},\hat{L}) > 2t \right) \nonumber\\ 
& \leq  
 \PP  \Big [ \underset{1 \leq K \leq n}{\max} \hspace{1mm} \underset{1 \leq L \leq K}{\max} \hspace{1mm} \underset{Z \in \mathcal{M}_{n,K}}{\max} \hspace{1mm} \underset{C \in \mathcal{M}_{K,L}}{\max}   
 ( |\langle \Xi(Z,C,K,L),H_{\tilde{u},\tilde{v}}(Z,C,K,L) \rangle|^2 -   
& F_2(n,K,L) )  > 2t \Big]  \label{eq:Del2_union} \\
& \leq  
 2 n K K^{n} L^{K}\exp \lfi -  F_2(n,K,L)/2 - t \rfi =  2 \exp(-t), \nonumber
\end{align}
where $F_2(n,K,L)=2 \ln n + 2(n+1) \ln K + 2K \ln L$.  
By Lemma~\ref{lem:Pi_orth}, one has 
\bes
\begin{aligned}
& \left\|\Pi_{\tilde{u}, \tilde{v}}\left( P_{*}(\hat{Z},\hat{C},\hat{K},\hat{L})\right)- P_{*} (\hat{Z},\hat{C},\hat{K},\hat{L}) \right\|_F^2 \leq \\
& \left\|\Pi_{\hat{u}, \hat{v}}\left( P_{*}(\hat{Z},\hat{C},\hat{K},\hat{L})\right)- P_{*} (\hat{Z},\hat{C},\hat{K},\hat{L}) \right\|_F^2 \leq 
\left\| \hat{P}  - P_{*}  \right\|_F^2.
\end{aligned}
\ees
Denote the set on which \eqref{eq:Del2_union} holds by $\Omega_{2}^C$, so that 
\be \label{eq:P_Omega2}
\PP(\Omega_{2}) \geq  1 - 2 \exp(-t).
\ee
Then inequalities \eqref{eq:Del2_sum}  and \eqref{eq:Del2_union} imply that, for any $\alpha_1>0$, $ t >0$ and 
any $\omega \in \Omega_{2}$,  one has
\be \label{eq:Del2_bound}
 |\Delta_2|  \leq  \alpha_1  \left\| \hat{P}  - P_{*}  \right\|_F^2 + 
1/\alpha_1 \, F_2(n,\hat{K},\hat{L}) + 2\,t/\alpha_1.
\ee


Now consider $\Delta_3$ defined in \eqref{delta3_main_est_error}
with components \eqref{kl_blockdelta3_main_est_error}. 
Note that matrices  $$\Pi_{\hat{u}, \hat{v}} (P_{*}^{(l,k)}(\hat{Z},\hat{C},\hat{K},\hat{L}) ) -
\Pi_{\tilde{u}, \tilde{v}} \left(P_{*}^{(l,k)}(\hat{Z},\hat{C},\hat{K},\hat{L})\right)$$
have rank at most two.
Use the fact that (see, e.g., Giraud  (2014), page 123)
\be \label{eq:Ky-Fan}
\begin{aligned}
\langle A, B \rangle  \leq \left\|A\right\|_{(2,r)} \left\|B\right\|_{(2,r)} \leq 2 \left\|A\right\|_{op} \left\|B\right\|_F, \quad r = \min \{\text{rank}(A), \text{rank}(B)\}.
\end{aligned}
\ee
Here $\left\|A\right\|_{(2,q)}$ is the Ky-Fan $(2,q)$ norm 
$$
\left\|A\right\|^2_{(2,q)} = \sum_{j=1}^q \sig_j^2(A) \leq \left\|A\right\|^2_F,
$$
where $\sig_j(A)$ are the singular values of $A$.
Applying inequality \eqref{eq:Ky-Fan} with $r=2$ and taking into account that for any matrix $A$ one has 
$\left\|A\right\|^2_{(2,2)} \leq 2 \left\|A\right\|_{op}^2$, derive
\bes
\begin{aligned}
 |\Delta_3^{(l,k)}| \leq  4 \left\| \Xi^{(l,k)}(\hat{Z},\hat{C},\hat{K},\hat{L})\right\|_{op}  \Big \|\Pi_{\hat{u}, \hat{v}} (P_{*}^{(l,k)}(\hat{Z},\hat{C},\hat{K},\hat{L}) 
- 
\Pi_{\tilde{u}, \tilde{v}} \left(P_{*}^{(l,k)}(\hat{Z},\hat{C},\hat{K},\hat{L})\right) \Big\|_F.
\end{aligned}
\ees
Then, for any $\alpha_2>0$, obtain 
\begin{align}
& |\Delta_3 | \leq  \di\sum_{l=1}^{\hat{L}} \sum_{k=1}^{\hat{K}} |\Delta_3^{(l,k)}|  
 \leq \frac{2}{\alpha_2} \di\sum_{l=1}^{\hat{L}} \sum_{k=1}^{\hat{K}} \left\| \Xi^{(l,k)}(\hat{Z},\hat{C},\hat{K},\hat{L})\right\|_{op}^2 +  \label{eq:Del3_sum} \\
& 2\alpha_2 \di\sum_{l=1}^{\hat{L}} \sum_{k=1}^{\hat{K}}  \Big\|  \Pi_{\hat{u}, \hat{v}} (P_{*}^{(l,k)}(\hat{Z},\hat{C},\hat{K},\hat{L})  \nonumber 
 - \Pi_{\tilde{u}, \tilde{v}} \left(P_{*}^{(l,k)}(\hat{Z},\hat{C},\hat{K},\hat{L})\right) \Big\|_F^2. \nonumber
\end{align} 

Note that, by Lemma~\ref{lem:Pi_orth}, 
\begin{align*}
& \left\|\Pi_{\hat{u}, \hat{v}} (P_{*}^{(l,k)}(\hat{Z},\hat{C},\hat{K},\hat{L})) -\Pi_{\tilde{u}\tilde{v}} \left(P_{*}^{(l,k)}(\hat{Z},\hat{C},\hat{K},\hat{L})\right) \right\|_F^2\\
\leq 
& 2 \left\|\Pi_{\hat{u}, \hat{v}} (P_{*}^{(l,k)}(\hat{Z},\hat{C},\hat{K},\hat{L})) - P_{*}^{(l,k)}(\hat{Z},\hat{C},\hat{K},\hat{L})\right\|^2_F + \\ 
& 2 \left\| \Pi_{\tilde{u}, \tilde{v}} (P_{*}^{(l,k)}(\hat{Z},\hat{C},\hat{K},\hat{L})) - P_{*}^{(l,k)}(\hat{Z},\hat{C},\hat{K},\hat{L})\right\|^2_F\\
\leq 
& 4 \left\|\Pi_{\hat{u}, \hat{v}} (P_{*}^{(l,k)}(\hat{Z},\hat{C},\hat{K},\hat{L})) - P_{*}^{(l,k)}(\hat{Z},\hat{C},\hat{K},\hat{L})\right\|^2_F \\
\leq 
& 4 \left\|\Pi_{\hat{u}, \hat{v}}(A^{(l,k)}(\hat{Z},\hat{C},\hat{K},\hat{L})) - P_{*}^{(l,k)}(\hat{Z},\hat{C},\hat{K},\hat{L}) \right\|_F^2  \\
& = 4 \left\| \hat{\Theta}^{(l,k)}(\hat{Z},\hat{C},\hat{K},\hat{L}) - P_{*}^{(l,k)}(\hat{Z},\hat{C},\hat{K},\hat{L}) \right\|_F^2  
\end{align*}
Therefore,
\begin{align}  
& \di\sum_{l=1}^{\hat{L}} \sum_{k=1}^{\hat{K}} 
 \left\|\Pi_{\hat{u}, \hat{v}} (P_{*}^{(l,k)}(\hat{Z},\hat{C},\hat{K},\hat{L})) -
\Pi_{\tilde{u},\tilde{v}} \left(P_{*}^{(l,k)}(\hat{Z},\hat{C},\hat{K},\hat{L})\right) \right\|_F^2 \leq  \nonumber\\
&  4  \left\|\hat{\Theta}(\hat{Z},\hat{C},\hat{K},\hat{L}) -P_{*}(\hat{Z},\hat{C},\hat{K},\hat{L})\right\|_F^2 
= 4 \left\| \hat{P}  - P_{*}  \right\|_F^2  \label{eq:Del3_part}
\end{align}
Combine  inequalities  \eqref{eq:Del3_sum} and \eqref{eq:Del3_part}  and recall  that 
$$\di\sum_{l=1}^{\hat{L}} \sum_{k=1}^{\hat{K}} \left\|\Xi^{(l,k)} (\hat{Z},\hat{C},\hat{K},\hat{L})\right\|_{op}^2 \leq  F_1(n,\hat{K},\hat{L}) + C_3 \, t$$ for 
$\omega \in \Omega_{1}$. Then,  for any $\alpha_2>0$  and  $\omega \in \Omega_{1}$, 
one has
\be \label{eq:Del3_bound}
  |\Delta_3| \leq 8 \alpha_2 \left\| \hat{P}  - P_{*}  \right\|_F^2 + 
2/\alpha_2 F_1(n,\hat{K},\hat{L}) + 2 C_3\, t/\alpha_2.
\ee

Now, let $\Omega =  \Omega_{1} \cap \Omega_{2}$. Then, \eqref{eq:P_Omega1} and \eqref{eq:P_Omega2}
imply that $\PP(\Omega) \geq 1 - 3 \exp(-t)$ and, for $\om \in \Omega$,
inequalities \eqref{eq:Del1_bound}, \eqref{eq:Del2_bound} and \eqref{eq:Del3_bound} simultaneously   hold.
Hence, by \eqref{eq:Del_sum}, derive that, for any $\om \in \Omega$,
\bes 
\begin{aligned}
& |\Delta| \leq (2 + 2/\alpha_2) F_1(n,\hat{K},\hat{L}))  + 1/\alpha_1\, F_2(n,\hat{K},\hat{L}) + \\
& (\alpha_1 + 8 \alpha_2) \left\| \hat{P}  - P_{*}  \right\|_F^2 + 2(C_3 + 1/\alpha_1 + C_3/\alpha_2)\, t.
\end{aligned}
\ees 
Combination of the last inequality and \eqref{eq:tot_err} yields that, for $\alpha_1 +  8\alpha_2 <1$ and any $\om \in \Omega$,
\begin{align*}
 (1-\alpha_1 - 8\alpha_2)\, \left\|\hat{P}  -P_{*}\right\|_F^2 \leq  
\left( 2 + \frac{2}{\alpha_2}\right) F_1(n,\hat{K},\hat{L})  + \\ 
\frac{1}{\alpha_1} F_2(n,\hat{K},\hat{L}) + \Pen(n,K_{*},L_{*}) - \Pen(n,\hat{K},\hat{L})\\
 + 2(C_3 + 1/\alpha_1 + C_3/\alpha_2)\, t   
\end{align*}
Setting $\Pen(n,K,L) =  (2 + 2/\alpha_2) F_1(n,K,L)   + 1/\alpha_1  F_2(n,K,L)$ and dividing 
by $(1-\alpha_1 - 8\alpha_2)$, obtain that 
\be \label{eq:tot_err_new}
\begin{aligned}
 \PP \lfi \left\|{\hat{P}  -P_{*}}\right\|_F^2 \leq  (1-\alpha_1 - 8\alpha_2)^{-1} \, \Pen(n,K_{*},L_{*}) + \tilde{C}\, t \rfi 
 \geq 1 - 3 e^{-t}
\end{aligned}
\ee
where
\be \label{eq:tildeC}
\tilde{C} =  2\, (1-\alpha_1 - 8\alpha_2)^{-1} \,(C_3 + 1/\alpha_1 + C_3/\alpha_2) 
\ee
%
%
%
%
%
Moreover, note that 
for $\xi = \|{\hat{P} -P_{*}}\|_F^2 -  (1 - \beta_1 -\beta_2)^{-1} \,  \Pen (n,K_{*},L_{*})$, 
one has 
$
\EE \|{\hat{P} - P_{*}}\|_F^2  = (1 - \beta_1 -\beta_2)^{-1} \,  \Pen (n,K_{*},L_{*})  + \EE \xi, 
$
where
\bes
\begin{aligned}
 \EE \xi \leq \int_0^{\infty} \PP(\xi > z) dz = \tilde{C}\int_0^{\infty} \PP(\xi >  \tilde{C}t) dt  
 \leq \tilde{C}\int_0^{\infty} 3 \, e^{-t}\, dt  = 3\tilde{C},
\end{aligned}
\ees
By rearranging and combining the terms, the penalty $\Pen(n,K,L)$ can be written in the form \eqref{eq:penalty} completing the proof.


\subsection{Proof  of detectability of clusters in Lemma~\ref{lem:detect}  }

Note that the left hand side of inequality \eqref{eq:detect} is equal to identical zero, so we need to show that, for any  
matrices  $Z \in \calM_{n, K}$ and $C \in \calM_{K, L}$ such that $Z$ and $C$ cannot be obtained from $Z_*$ and $C_*$ 
by   permutations of columns, the right-hand side of  \eqref{eq:detect} is  greater than zero. Consider matrix $G_* =Z_* C_*$.
It is easy to check that $G_* \in \calM_{n, L}$  is the clustering matrix that partitions $n$ nodes into $L$ meta-communities.
We first show that $G_*$ coincides with  $G = ZC$   up to permutation of columns. Subsequently, we shall show that communities 
in each meta-community are identifiable.

Consider a block $P_*^{(l,k)}(Z_*,C_*)$ of matrix $P_*$. Let $c(k) =m$. 
Let $j_1, \ldots, j_{K_l}$ be the indices of the communities in meta-community $l$, so that $c(j_t) = l$.
Then, $P_*^{(l,k)}(Z_*,C_*)$ is a rank one matrix which is a product of the vector obtained by the vertical concatenation 
of vectors $B_{j_t,k} h^{(j_t,m)}$, $t=1, \ldots, K_l$ and $(h^{(k,l)})^T$. Assume that a node $\tj$ such that $z(\tj)= \tk$
and $c(\tk) = \tl \neq l$ was erroneously placed into meta-community $l$ instead of $\tl$. This is equivalent to adding a row 
$c_{0,\tj} \, (h^{(k,\tl)})^T$ to matrix $P_*^{(l,k)}(Z_*,C_*)$ where $c_{0,\tj} = B_{\tk,k} \, h_{\tj}^{(\tk,m)}$. If 
$c_{0,\tj} \neq 0$, then the  resulting matrix will be of rank at least two, since vectors $h^{(k,\tl)}$ and $h^{(k,l)}$ are linearly independent 
for $\tl \neq l$ by Assumption A1. If $c_{0,\tj} = B_{\tk,k} \, h_{\tj}^{(\tk,m)}= 0$, find $k$ such that 
$c_{0,\tj} = B_{\tk,k} \, h_{\tj}^{(\tk,m)}\neq 0$  and then repeat the previous argument for the matrix $P_*^{(l,k)}(Z_*,C_*)$ for that value of $k$.
Note that such   $k$ exists since, otherwise, row $\tj$ would be identically equal to zero and, hence,
node $\tj$ would be disconnected from the network.

Therefore, meta-communities are detectable. To prove that communities within meta-communities are identifiable, consider 
diagonal meta-blocks $\tilde{P}^{(l,l)} (Z_*,C_*)$ in \eqref{eq:P_tilde_blocks}. It follows from   Interlace Theorem for
eigenvalues and Assumption A1 that $\lam_{\min} (B^{(l,l)}) \geq \lambda_{\min}(B) \ge \lambda_0 > 0$,
and therefore columns  of matrix $B^{(l,l)}$ are linearly independent.
Now, consider again matrix $P_*^{(l,k)}(Z_*,C_*)$, where  $m = c(k)=l$. Recall that columns of this matrix are multiples of vector $\tp^{(k)}$  
obtained by the vertical concatenation of vectors $B_{j_t,k} h^{(j_t, l)}$, $t=1, \ldots, K_l$. Now, assume that node $\tj$ with 
$z(\tj) =\tk \neq k$, $c(\tk) = l$, is erroneously added to the community $k$. Then, the corresponding column that is added to matrix 
$\tilde{P}^{(l,l)} (Z_*,C_*)$ is obtained by the vertical concatenation of vectors $B_{j_t, \tk} h^{(j_t, l)}$, $t=1, \ldots, K_l$.
This vector is linearly independent from  $\tp^{(k)}$ due to linear independence of columns of matrix  $B^{(l,l)}$.
Then, the rank of the resulting matrix will be at least two and the right hand side of the inequality \eqref{eq:detect}  will be positive. 
This completes the proof. 


%
 
\medskip

\medskip


\subsection{Supplementary statements and their proofs}
\label{sec:Suppl_statements}

\begin{lemma}  \label{lem:log_size_delta_net}
The logarithm of the cardinality  of a $\delta$-net on the space of  non-symmetric DCBMs 
of size $n_1 \times n_2$ with $K_1 \times K_2$ blocks is
\bes
(K_1 K_2 + n_1 + n_2) \ln \left (\frac{9}{\delta} \right ) + 
\lkr K_1 K_2 + \frac{n_1 + n_2}{2}\rkr \ln(n_1 n_2)
\ees
\end{lemma}


\noindent{\bf Proof.}
Let $Z_1$ and $Z_2$ be fixed. Then by rearranging $\Theta$, rewrite it as $\Theta=Q_1 B Q_2^T$, where $B$ and $Q_i$, $i=1,2$, have the sizes $K_1 \times K_2$ and $n_i \times K_i$, respectively.
Here, $Q_i$ is of the form
\be    \label{eq:Qi_matrices}  
\begin{aligned}
 Q_i=
  \begin{bmatrix}
     q_{i,1} & \boldsymbol{0} & \cdots  & \boldsymbol{0}\\
     \boldsymbol{0} & q_{i,2} & \cdots  & \boldsymbol{0}\\
    \vdots & \vdots & \cdots & \vdots\\
    \boldsymbol{0} & \boldsymbol{0} & \cdots  & q_{i,K_i}\\
  \end{bmatrix},   
\end{aligned}
\ee 
We re-scale components of matrices $Q_1$, $Q_2$ and $B$, so that 
vectors $q_{i,j} \in \RR_+^{n_{i,j}}$, $j=1,\cdots,K_i,$ $i=1,2$,
have unit norms $\left\|q_{i,j}\right\|_2=1$, and $\sum_{j = 1}^{K_i} n_{i,j}=n_i$. 
Let $\Theta^{(k_1, k_2)} \in \RR^ {n_{1,k_1} \times n_{2,k_2}}$ 
be the $(k_1,k_2)$-th block of $\Theta$. Then,
\bes
\Theta^{(k_1, k_2)} = B_{k_1, k_2} q_{1,k_1} q_{2,k_2}^T 
\ees
and
\bes
\begin{aligned}
\left\|\Theta^{(k_1, k_2)}\right\|_F^2 = B_{k_1, k_2}^2 \left\|q_{1,k_1}\right\|_2^2 \left\|q_{2,k_2}\right\|_2^2 = 
B_{k_1, k_2}^2 \leq n_{k_1} \cdot n_{k_2},
\end{aligned}
\ees
due to $\left\|a b^T\right\|_F^2 \leq  \left\|a\right\|_2^2
  \left\|b\right\|_2^2$ (for any vectors $a$ and $b$) and
  $\left\|\Theta\right\|_{\infty} \leq 1$. Hence, $B_{k_1, k_2} \leq \sqrt{n_{k_1} \cdot n_{k_2}} \leq \sqrt{n_1 \cdot n_2}$.

Let $\mathscr{D}_1(\delta_1)$, $\mathscr{D}_2(\delta_2)$, and $\mathscr{D}_B(\delta_B)$ be the $\delta_1$, $\delta_2$,  and $\delta_B$ nets  for $Q_1$, $Q_2$, and $B$, respectively. 
The nets $\mathscr{D}_i(\delta_i)$ are essentially constructed for $K_i$ 
vectors of length 1 in $\RR^{n_{i,j}}$, hence, by \cite{pollard1990empirical}
\bes
\begin{aligned}
\card( \mathscr{D}_i(\delta_i)) \leq  \Pi_{j=1}^{K_i} \left ( 3/\delta_i  \right )^{n_{i,j}} = \left (  3/\delta_i \right )^{n_{i}}, \ i=1,2.
\end{aligned}
\ees
Let $b=\vect{(B)}$. Then, $b \in \RR ^ {K_1 K_2}$ and $\left\|b\right\| \leq \sqrt{n_1 n_2}$ since
\bes
\begin{aligned}
\left\|b\right\|^2=\left\|B\right\|_F^2 = \sum_{k_1,k_2} B_{k_1,k_2}^2 = \sum_{k_1,k_2} n_{k_1}n_{k_2} = n_1 n_2.
\end{aligned}
\ees
Therefore,
\bes
\begin{aligned}
\card( \mathscr{D}_B(\delta_B)) \leq   
\left ( \frac{3 n_1 n_2}{\delta_B} \right )^{K_1 K_2} 
\end{aligned}
\ees
Now, let us check what values of $\delta_1$, $\delta_2$, and $\delta_B$ result in 
a $\delta$-net. Let $\Theta=Q_1 B Q_2^T$ and $\tilde{\Theta}=\tilde{Q}_1 \tilde{B} \tilde{Q}_2^T$. Then
\bes
\begin{aligned}
&\left\|\tilde{\Theta} - \Theta\right\|_F = \left\|\tilde{Q}_1 \tilde{B} \tilde{Q}_2^T - Q_1 B Q_2^T\right\|_F \leq \\ &\left\|(\tilde{Q}_1 - Q_1) \tilde{B} \tilde{Q}_2^T\right\|_F + 
\left\|Q_1 (\tilde{B}-B) \tilde{Q}_2^T\right\|_F + 
\left\|Q_1 B (\tilde{Q}_2-Q_2)^T\right\|_F
\end{aligned}
\ees
Note that 
\bes
\begin{aligned}
\left\|A_1 A_2\right\|_F \leq \min \lkr \left\|A_1\right\|_F \left\|A_2\right\|_{op}, 
\left\|A_1\right\|_{op} \left\|A_2\right\|_F \rkr
\end{aligned}
\ees
for any matrices $A_1$ and $A_2$, and that also 
\bes
\begin{aligned}
Q_i^T Q_i =\diag{\left(\left\|q_{i,1}\right\|^2, \cdots, \left\|q_{i,K_i}\right\|^2\right)} = I_{K_i}, \hspace{1mm} i=1,2.
\end{aligned}
\ees
Hence
\bes
\begin{aligned}
\norm{Q_i}_{op}=1; \hspace{1mm} \norm{Q_i}_F=\sqrt{K_i}, \hspace{1mm} i=1,2.
\end{aligned}
\ees
Similarly, if $\tilde{Q}_i , Q_i \in \mathscr{D}_i(\delta_i)$, then 
\bes
\begin{aligned}
& (\tilde{Q}_i - Q_i)^T (\tilde{Q}_i - Q_i) = 
& \diag{\left(\left\|\tilde{q}_{i,1} - q_{i,1}\right\|^2, \cdots, \left\|\tilde{q}_{i,K_i} - q_{i,K_i}\right\|^2\right)}
\end{aligned}
\ees
Thus
\bes
\begin{aligned}
\left\|\tilde{Q}_i - Q_i\right\|_{op} = \delta_i; \hspace{2mm} \left\|\tilde{Q}_i - Q_i\right\|_F \leq \sqrt{K_i}  \delta_i, \hspace{1mm} i=1,2.
\end{aligned}
\ees
Also, for $i=1,2$
\bes
\begin{aligned}
\Tr (B^T Q_i^T Q_i B) = \left\|Q_i B\right\|_F^2 = \left\|B\right\|_F^2 = n_1 n_2.
\end{aligned}
\ees
Hence,
\bes
\begin{aligned}
& \left\|\tilde{\Theta} - \Theta\right\|_F \leq \left\|\tilde{Q}_1 - Q_1\right\|_{op} \left\|\tilde{B} \tilde{Q}_2^T\right\|_F  \\
& +  \left\|Q_1 B\right\|_F \left\|\tilde{Q}_2 - Q_2\right\|_{op} +  \left\|Q_1\right\|_{op} \left\|\tilde{B} - B\right\|_F \left\|\tilde{Q}_2 \right\|_{op} \\
& = 
 (\delta_1 + \delta_2) \sqrt{n_1 n_2} + \delta_B \leq \delta
\end{aligned}
\ees
Set $\delta_B = \frac{\delta}{3}$ and $\delta_1 = \delta_2 = \frac{\delta}{3 \sqrt{n_1 n_2}}$. Then
\bes
\begin{aligned}
&\card( \mathscr{D}_B(\delta_B)) =  \left ( \frac{9 n_1 n_2}{\delta} \right )^{K_1 K_2}, \\
&\card( \mathscr{D}_i(\delta_i)) =  \left ( \frac{9 \sqrt{n_1 n_2}}{\delta} \right )^{n_i},
\end{aligned}
\ees
which completes the proof. 
\\

\medskip

\begin{lemma}  \label{lem:size_epsilon_net}
Consider the set of  matrices $P$ which can be  transformed by a permutation matrix $\mathscr{P}_{Z,C}$ into a block matrix $\Theta \in \Im(n,K,L)$  where $\Im(n,K,L)$ is 
defined in \eqref{eq:opt_cond}.
\ignore{
with $L^2$ meta-blocks where each meta-block follows the DCBM model and the sizes and the number of blocks in each meta-block are defined by matrices $Z \in \calM_{n,K}$ and $C \in \calM_{K,L}$ and $K^2$ is the total number of blocks. Denote
\be
\begin{align} \label{eq:set_M}
\mathscr{M}(n,K,L)= \left \{ P: \hspace{1mm} P = \mathscr{P}_{Z,C} \Theta \mathscr{P}_{Z,C}^T , \hspace{1mm} Z \in \calM_{n,K}; \hspace{1mm} \right \\
\left C \in \calM_{K,L}; \hspace{1mm} \Theta = \cup_{l_1, l_2 =1}^{L} \Theta^{(l_1,l_2)} ; \hspace{1mm} P \in [0,1]^{n \times n}    \right \} 
\end{align}
\ee
} 
Let $\mathscr{Y}(\epsilon, n, K, L)$ be an $\epsilon$-net on 
the set $\Im(n,K,L)$ and $| \mathscr{Y}(\epsilon, n, K, L) |$ 
be its cardinality. Then, for any $K$ and $L$, 
$1 \leq K \leq n,$ $1 \leq L \leq K,$ one has
\be
\begin{aligned} \label{eq:size_epsilon_net}
| \mathscr{Y}(\epsilon, n, K, L) | \leq n \ln K + K \ln L +
(K^2 + 2 n L ) \ln \left( \frac{9 n L}{\epsilon} \right)
\end{aligned}
\ee
\end{lemma}


\noindent{\bf Proof.}
First construct   nets on the set of matrices $Z$ and $C$ with the respective cardinalities  $K^n$ and  $L^K$. After that, validity of the lemma follows from   Lemma~\ref{lem:log_size_delta_net}.
\\


\medskip

\begin{lemma}  \label{lem:upper_bound_klopp}
Let $C_0^2 = 3009$, $C_2 =1$, $s_0>0$  be an arbitrary constant and
$\Omega_{K,L}$  be defined in \eqref{eq:Omkl}. 
Then,
\bes
\begin{aligned} 
\PP \bigg \{ \underset{\hat{P} \in  \Omega_{K,L}}{\rm{sup}}  \Big [ 2 \langle \Xi , \hat{P} - P_* \rangle - \frac{1}{2} \left\|\hat{P}  -P_{*}\right\|_F^2 - 
 2 \Delta (n, K, L) \Big ] \geq 0 \bigg \} \leq 
 \log_2 n \cdot \exp{\big( -n \cdot 2^{2s_0-7} \big)}
\end{aligned}
\ees
where  $\Delta(n,K,L)$ is defined in \eqref{eq:DelnKL}.
\end{lemma}


\noindent{\bf Proof.}
Consider sets
\bes
\begin{aligned} 
  &\chi_s (K,L)   = \Big \{ \exists Z,C:\ 
  P(Z,C) \in \Im (n,K,L);  \\
 &  C_0 2^s \sqrt{\tau (n, K_0, L_0)} \leq  
 \|P  -P_{*}\|_F \leq C_0 2^{s+1} \sqrt{\tau (n, K_0, L_0)} \Big \},
\end{aligned}
\ees
and
\bes
\begin{aligned} 
\calJ_s (K,L)   = \Big \{ & \exists Z,C:\ 
  P(Z,C) \in \Im (n,K,L); 
& \|P  -P_{*}\|_F \leq    C_0 2^{s} \sqrt{\tau (n, K_0, L_0)} \Big \}
\end{aligned}
\ees
Note that the set $\Omega$ can be partitioned as
\bes
\Omega = \underset{K,L} {\bigcup} \Omega_{K,L}
\ees
where $\Om_{K,L}$ are defined in \eqref{eq:Omkl}. 
Then
\bes
\begin{aligned} 
& \PP \bigg \{ \underset{\hat{P} \in \Omega_{K,L}}{\rm{sup}} \Big [ \langle \Xi , \hat{P} - P_* \rangle - \frac{1}{4} \left\|\hat{P} -P_*\right\|_F^2 - 
\Delta (n, K, L) \Big ] \geq 0 \bigg \} \leq \\
& \sum_{s= s_0}^{s_{\max}} \PP \bigg \{ \underset{\hat{P} \in \chi_s (K,L)}{\text{sup}} \Big [ \langle \Xi , \hat{P} - P_* \rangle - \frac{1}{4} \left\|\hat{P} -P_*\right\|_F^2 - 
  \Delta (n, K, L) \Big ] \geq 0 \bigg \} \leq\\
& \sum_{s= s_0}^{s_{\max}} \PP \bigg \{ \underset{\hat{P} \in \chi_s (K,L)}{\text{sup}}  \langle \Xi , \hat{P} - P_* \rangle   \geq  C_0^2 2^{2 s - 2} \tau (n, K_0, L_0) +  
 \Delta (n, K, L)  \bigg \} \leq\\
& \sum_{s= s_0}^{s_{\max}} \PP \bigg \{ \underset{\hat{P} \in \calJ_{s+1} (K,L)}{\text{sup}}  \langle \Xi , \hat{P} - P_* \rangle  \geq  C_0^2 2^{2 s - 2} \tau (n, K_0, L_0) + 
 \Delta (n, K, L)  \bigg \}
\end{aligned}
\ees
Here, $s_{max} \leq \log_2 n $ since $\left\|\hat{P} -P_*\right\|_F \leq n$.

Construct a 1-net  $\mathscr{Y}_s ( n, K, L)$ on the set of matrices in $\calJ_{s+1} (K,L)$ and observe that, for any $\hat{P} \in \calJ_s (K,L)$, there exists $\tilde{P} \in \mathscr{Y}_s ( n, K, L)$ such that $\| \hat{P} - \tilde{P} \|_F \leq 1$. Then,
\bes
\begin{aligned} 
& \underset{\hat{P} \in \mathscr{Y}_{s+1} ( n, K, L) }{\text{sup}} \langle \Xi,
\hat{P} - P_* \rangle \leq \\
&\underset{\tilde{P} \in \mathscr{Y}_{s} ( n, K, L) }{\text{max}} \big [\langle \Xi,
\tilde{P} - P_* \rangle  + \langle \Xi , \hat{P} - \tilde{P} \rangle \big ] \leq \\
&\underset{\tilde{P} \in \mathscr{Y}_{s} ( n, K, L) }{\text{max}} \langle \Xi, \tilde{P} - P_* \rangle  + n
\end{aligned}
\ees
Hence,
\bes
\begin{aligned} 
& \PP \bigg \{ \underset{\hat{P} \in \Omega_{K,L}}{\text{sup}} \Big [ \langle \Xi,
\hat{P} - P_* \rangle   - \frac{1}{4} \left\|\hat{P} -P_*\right\|_F^2     
   - \Delta (n, K, L) \Big ] \geq 0 \bigg \} \leq \\
& \sum_{s= s_0}^{s_{\max}} \PP \bigg \{ \underset{\tilde{P} \in \mathscr{Y}_{s}
(n, K, L)}{\text{max}}  \langle \Xi , \tilde{P} - P_* \rangle   \geq  
C_0^2 2^{2 s - 2} \tau (n, K_0, L_0) 
 + \Delta (n, K, L) - n \bigg \} \leq\\
& \sum_{s= s_0}^{s_{\max}} \sum_{\tilde{P} \in \mathscr{Y}_{s} ( n, K, L)}^{} 
\PP \bigg \{ \langle \Xi , \tilde{P} - P_* \rangle   \geq  C_0^2 2^{2 s - 2} 
\tau (n, K_0, L_0)   
 +\Delta (n, K, L) - n \bigg \} 
\end{aligned}
\ees
Below we shall use the following version of Bernstein inequality (see, e.g., \cite{klopp2019}, Lemma~26):
if $\Xi$ is a matrix of independent Bernoulli errors and $G$ is an arbitrary matrix
of the same size, then for any $t>0$ one has
\be \label{eq:Bernstein}
\PP \lfi \langle \Xi,G \rangle > t \rfi \leq \max \lkr 
e^{ - \frac{t^2}{4 \|G\|^2_F}}, e^{ - \frac{3t}{4\|G\|_{\infty}}} \rkr.
\ee
We apply \eqref{eq:Bernstein} with $G = \tilde{P} - P_*$  and 
\be \label{eq:tval}
  t = C_0^2 \Big [2^{2 s - 2} \tau (n, K_0, L_0) + C_2 \tau (n, K, L) \Big ]. 
\ee 
Then, $\|G\|_{\infty} =1$  and $\| G \|^2 \leq C_0^2 2^{2 s + 2} \tau (n, K_0, L_0)$
due to
$\tilde{P} \in \mathscr{Y}_{s} ( n, K, L) \subseteq \calJ _{s+1} (K,L)$.

Denote 
\be \label{eq:dkls}
d_{K,L}^{(s)} =  
\max \lfi   e^{ - \frac{t^2}{4 C_0^2 2^{2s+2}\, \tau (n, K_0, L_0)}},
e^{- \frac{3 t}{4}} \rfi
\ee  
\be \label{eq:dkl}
d_{K,L} =  
\sum_{s= s_0}^{s_{\max}} d_{K,L}^{(s)} \cdot \exp \lfi \tau(n,K,L) \rfi
\ee
Obtain
\be \label{eq:dkl_bound}
\begin{aligned} 
\PP \bigg \{ \underset{\hat{P} \in \Omega_{K,L}}{\rm{sup}} & \Big [ \langle \Xi , \hat{P} - P_* \rangle - \frac{1}{4} \left\|\hat{P} -P_*\right\|_F^2 
 -  \Delta (n, K, L) \Big ] \geq 0 \bigg \} \leq d_{K,L}
\end{aligned}
\ee 
Observe that
\bes
\begin{aligned} 
\exp{ \Big \{ - \frac{t^2}{4 C_0^2 2^{2s+2} \tau (n, K_0, L_0)} \Big \} } \geq   \exp{ \Big \{ - \frac{3 t}{4} \Big \} } 
\end{aligned}
\ees
is equivalent to $t \leq 3 C_0^2 2^{2s+2} \tau (n, K_0, L_0)$
which can be rewritten as 
\be \label{eq:case_1}
C_2 \tau (n, K, L) \leq 47 \cdot 2^{2s-2} \tau (n, K_0, L_0)
\ee
Now, consider two cases: when \eqref{eq:case_1} holds and when it does not.

{\it Case 1:} If (\ref{eq:case_1}) holds, then
\bes
\begin{aligned} 
 d_{K,L}^{(s)} \leq  
%
\exp \bigg \{ - C_0^2  \Big [ 2^{2s-8} \tau (n, K_0, L_0) + \frac{C_2^2 \tau ^2 (n, K, L) }{ 2^{2s+4} \tau(n, K_0, L_0)} \Big ] \bigg \},  
\end{aligned}
\ees
so that 
\bes
\begin{aligned} 
& d_{K,L}^{(s)}\, \exp{\big \{   \tau (n, K, L) \big \} }  \leq  \\
&\exp \bigg \{ -  \Big [ C_0^2  2^{2s-8} \tau (n, K_0, L_0) -  
  \frac{47 \cdot 2^{2s-2} }{C_2}  \tau(n, K_0, L_0) \Big ] \bigg \} \leq \\
&\exp \bigg \{ - \tau(n, K_0, L_0) \cdot 2^{2 s_0 -8 } \Big [ C_0^2 -  
  \frac{47 \cdot 64 }{C_2}  \Big ] \bigg \} .
\end{aligned}
\ees
Thus, it follows from \eqref{eq:dkls} and  \eqref{eq:dkl}  that
\be \label{eq:dkl1}
d_{K,L}   \leq \log_2 n \cdot \exp \lfi -  \tau(n, K_0, L_0) 2^{2 s_0 -8} 
\tilde{C} \rfi
\ee 
where $\tilde{C}   = ( C_0^2 C_2 - 47 \cdot 64)/C_2$,  
provided  $C_0 C_2 \geq 47 \cdot 64$.

{\it Case 2:} If  (\ref{eq:case_1}) does not hold, then  
\bes
\begin{aligned} 
& d_{K,L}^{(s)} \leq \\
& \exp \bigg \{ - \frac{3 C_0^2}{4} \Big [  2^{2s-2} \tau (n, K_0, L_0) + C_2 \tau (n, K, L) \Big ] \bigg \} \leq \\
&   \exp \bigg \{ - \tau (n, K, L)  - \tau (n, K, L) \big ( \frac{3 C_0^2 C_2}{4} -1 \big ) \bigg \}  
\end{aligned}
\ees
Hence, if $3 C_0^2 C_2 > 4$, then
\be \label{eq:dkl2}
\begin{aligned} 
d_{K,L} \leq \log_2 n \cdot \exp \bigg \{ - \tau(n, K, L)   
\Big (  \frac{3 C_0^2 C_2 - 4}{4} \Big ) \bigg \}.
\end{aligned}
\ee 
Combine \eqref{eq:dkl1} and \eqref{eq:dkl2} and observe that for 
$C_2 =1$ and $C_0^2=47 \cdot 64 +1 =3009$  inequalities 
$C_0 C_2 \geq 47 \cdot 64$ and $3 C_0^2 C_2 > 4$ hold.
Then, due to $\tau (n, K, L) \geq 2 n$, for any $(K,L)$
\bes
\begin{aligned} 
d_{K,L} \leq \log_2 n \cdot \exp \Big \{ - 2 n \cdot 2^{2 s_0 -8} \Big \}, 
\end{aligned}
\ees
so that validity of the lemma follows from \eqref{eq:dkl_bound}.
\ignore{
\bes
\begin{align} 
\PP \bigg \{ \underset{\hat{P} \in \Omega_{K,L}}{\text{sup}} \Big [ \langle \Xi , \hat{P} - P_* \rangle - \frac{1}{4} \norm{\hat{P} -P_*}_F^2 - \\
\big (  n + C_0^2 C_2 \tau (n, K, L) \big )\Big ] \geq 0 \bigg \} \leq \\
\log_2 n \cdot \exp \Big \{ -  n \cdot 2^{2 s_0 -7} \Big \}. 
\end{align}
\ees
} 
\\



\medskip

\begin{lemma}  \label{lem:lowrank_approx}
For any matrices $A, B \in \RR^{m \times n}$ and any unit vectors $u \in \RR^m$ and $v \in \RR^n$, let 
\be \label{eq:Pi_uv}
\Pi_{u,v}(A) =  (uu^T)A(vv^T)   
\ee 
denote the projection of matrix $A$ on the vectors $(u,v)$. Then,
\be \label{eq:Inner_Prod_Pi_uv}
\langle \Pi_{u,v}(B), A - \Pi_{u,v}(A) \rangle =0.
\ee
Furthermore, if we let $\hat{u}$ and $\hat{v}$ be the singular vectors of matrix $A$ corresponding to its largest singular value $\sig$, the best rank one approximation of $A$ is given by 
\be \label{eq:Pi_uv_A}
\Pi_{\hat{u},\hat{v}}(A) = (\hat{u} \hat{u}^T)A(\hat{v}\hat{v}^T)=  \sigma \hat{u}  \hat{v} ^T.
\ee 
\end{lemma}


\medskip

\begin{lemma}  \label{lem:Pi_orth}
Let $(\hat{u}, \hat{v})$ and $(u,v)$ denote the pairs of singular vectors of matrices $A$ and $P$, 
respectively, corresponding to their largest singular values. Then,
\be \label{eq:Pi_orth}
\left\|\Pi_{u,v} (P) - P\right\|_F \leq \left\|\Pi_{\hat{u},\hat{v}} (P) - P\right\|_F  \leq \left\|\Pi_{\hat{u},\hat{v}} (A) - P\right\|_F
\ee 
where $\Pi_{u,v} (\cdot)$ is defined in \eqref{eq:Pi_uv}.
\end{lemma}


\noindent{\bf Proof.}
See \cite{2019arXiv190200431N} for the proof.
\ignore{
The first inequality in \eqref{eq:Pi_orth}  is true because $\Pi_{u,v} (P)$ is the best rank one approximation of $P$. Now let $A=P+\Xi$. Then 
 \bes
\begin{aligned}
 \left\|\Pi_{\hat{u},\hat{v}} (A) - P\right\|_F^2  
= \left\|\Pi_{\hat{u},\hat{v}} (P) - P + \Pi_{\hat{u},\hat{v}} (\Xi)\right\|_F^2 =  \left\|\Pi_{\hat{u},\hat{v}} (P) - P\right\|_F^2 +\left \|\Pi_{\hat{u},\hat{v}} (\Xi)\right\|_F^2
\end{aligned}
\ees 
which leads to the second inequality in  \eqref{eq:Pi_orth}.
}

\ignore{ 
begin{lemma} \label{lem:prob_bound_error}
Let elements of matrix $\Xi \in (-1,1)^{n \times n}$ be independent Bernoulli 
Then, for any $x >0$  
\be \label{eq:lem3}
\PP \left\{ \sum_{k,l =1}^K \norm{\Xi^{(k,l)}}_{op}^2  \leq  C_1 nK + C_2 K^2 \ln (ne) + C_3x \right\}   \geq  1 - \exp (-x),
\ee 
where $C_1,C_2$ and $C_3 $ are absolute constants independent of $n$ and $K$. 
\end{lemma}
} 


\medskip

\begin{lemma} \label{lem:prob_bound_error}
Let elements of matrix $\Xi \in (-1,1)^{n \times n}$ be independent Bernoulli errors and matrix   $ \Xi$ be partitioned into $KL$ sub-matrices $\Xi^{(l,k)}$,  $l = 1, \cdots, L,$  $k = 1, \cdots, K$.
Then, for any $x >0$  
\be \label{eq:lem3}
\begin{aligned}
\PP \left\{ \sum_{l =1}^L\sum_{k =1}^K \left\|\Xi^{(l,k)}\right\|_{op}^2 \leq  C_1 nK + C_2 K^2 \ln (ne) + C_3x \right\} 
\geq  1 - \exp (-x),
\end{aligned}
\ee 
where $C_1,C_2$ and $C_3 $ are absolute constants independent of $n$,$K$, and $L$. 
\end{lemma}

\noindent{\bf Proof.} 
See \cite{2019arXiv190200431N} for the proof.



\medskip

\begin{lemma}\label{lem:prob_error_bound_main}
For any $ t > 0$,
\be \label{rr:prob_bound_Xi_case1} 
\begin{aligned}
\PP \left\{   \di\sum_{l = 1}^{\hat{L}}\sum_{k = 1}^{\hat{K}} \left\|\Xi^{(l,k)}(\hat{Z},\hat{C},\hat{K},\hat{L})\right\|_{op}^2  -F_1(n,\hat{K},\hat{L}) \leq   C_3 t \right\} 
\geq  1 - \exp{(-t)}.
\end{aligned}
\ee
where $F_1(n,K,L)=  C_1n K + C_2 K^2 \ln (ne) +C_3( \ln n + (n+1) \ln K + K \ln L).$
\end{lemma}


\noindent{\bf Proof. }
See \cite{2019arXiv190200431N} for the proof.

\bibliographystyle{spbasic}      
\bibliography{NBM}



\end{document}